\documentclass{article}

\usepackage[sglblindworkshop]{neurips_2026}

\usepackage[utf8]{inputenc} 
\usepackage[T1]{fontenc}    
\usepackage{hyperref}       
\usepackage{url}            
\usepackage{booktabs}       
\usepackage{pifont}
\usepackage{amsfonts}       
\usepackage{nicefrac}       
\usepackage{microtype}      
\usepackage{xcolor}         
\usepackage{graphicx}
\usepackage{amsmath}
\usepackage{amssymb}
\usepackage{makecell} 
\usepackage{multirow}
\usepackage{multicol}
\usepackage{wrapfig}
\usepackage{subcaption}
\usepackage{algorithm}
\usepackage{algpseudocode}
\usepackage[table]{xcolor}
\usepackage{hyperref}
\hypersetup{hidelinks}
\usepackage{xspace}        
\usepackage{footmisc}

\newcommand{\method}{RVEDiT\xspace}
\title{Reasoning to Align: Implicit Reasoning in Diffusion Transformers for Video Editing}

\author{%
  Yan Li$^{\dagger}$\footnotemark[1] \\
   \And
   Lin Liu$^{\ddagger}$ \\
   \And
   Xiaopeng Zhang$^{\ddagger}$ \\
   \And
   Qi Tian$^{\ddagger}$ \\
   \AND
   \vspace{-24pt}
   $^{\dagger}$\text{The Hongkong University of Science and Technology}
   \And
   $^{\ddagger}$\text{Huawei Inc.}
}

\begin{document}
\nolinenumbers

\footnotetext[1]{For any inquiry, please reach out to: ylitz@connect.ust.hk}
\maketitle

\begin{abstract}
Instruction-based video editing requires transforming a source video according to a natural-language instruction while preserving irrelevant content and remaining temporally coherent. We argue that existing Diffusion Transformer (DiT) editors struggle with this task for two structural reasons. First, conditioning signals are fed \emph{undifferentiated} into all transformer blocks, forcing a single token stream to encode both global editing intent and fine-grained visual evidence. Second, the cross-attention patterns that govern the edit are supervised only \emph{indirectly} through pixel-level reconstruction, leaving the model's internal reasoning process underconstrained.
To address both limitations, we propose \textbf{\method}, an implicit \textbf{R}easoning \textbf{V}ideo \textbf{E}diting \textbf{DiT} framework built around two complementary components. The first, \emph{Granularity-Routed Token Conditioning}, introduces learnable editing tokens distilled from a multimodal LLM and routes them to shallow blocks, while reserving native visual and textual tokens for deeper blocks, thereby inducing a coarse-to-fine editing process inside the backbone. The second, \emph{Reference-Anchored Attention Alignment}, employs a parameter-sharing reference branch during training and maximizes the mutual information between the attention features of the editing and reference branches, regularizing the model's internal reasoning without incurring any additional inference cost. Experiments on standard instruction-based video editing benchmarks show that \method consistently outperforms state-of-the-art baselines, with particularly strong gains on localized and compositional edits.
\end{abstract}

\section{Introduction} \label{sec:intro}
Video has become the dominant medium of visual communication, yet editing it remains a labor-intensive craft that demands expertise in timeline-based software, keyframe manipulation, and frame-level compositing.
Recent progress in large language models (LLMs)~\citep{brown2020language,achiam2023gpt} and text-conditioned diffusion models~\citep{ho2020denoising,rombach2022high} has opened a compelling alternative: \emph{instruction-based video editing}, in which users describe a desired modification in natural language (e.g., ``replace the dog with a corgi'' or ``make it look like a Van~Gogh painting'') and the system produces the edited video automatically, promising to make high-quality editing accessible to a far broader population of creators, educators, and everyday users~\cite{cong2025viva}.
As in instruction-based image editing~\citep{brooks2023instructpix2pix,hertz2022prompt}, video editing is not merely a text-to-pixel mapping problem, but a \emph{latent reasoning problem}: given a free-form instruction and a source video, the model must infer \emph{which} spatiotemporal regions to modify, \emph{what} transformation to apply, and \emph{what} must remain invariant across frames~\citep{he2025openve}.
A natural response is to externalize this reasoning by letting a multimodal LLM emit explicit plans, masks, or programs for the diffusion model to execute~\citep{yu2025veggie,yang2026omni,zhang2026meta}, but this outsources correctness to an external MLLM whose own biases and hallucinations propagate into the plan and cannot be corrected downstream, so the editor inherits whatever mistakes the MLLM makes.
Existing systems largely sidestep the issue instead: UNet-based editors~\citep{wu2023tune,qi2023fatezero,liu2024video} either inflate image-level backbones along time~\citep{brooks2023instructpix2pix} or scale triplet corpora atop unchanged architectures~\citep{he2025openve,baiditto,zi2025se,zhang2025region}, treating the backbone as a black box, and these recipes transfer poorly to Diffusion Transformers (DiTs)~\citep{peebles2023scalable}, whose global token mixing departs sharply from the locality bias of UNets.
It therefore remains an open question how a DiT backbone can be endowed with the capacity to perform this implicit reasoning natively, so that instruction faithfulness, source preservation, and temporal coherence emerge from a single, end-to-end generative pass.

We argue that current DiT pipelines fail to support such implicit reasoning for two structural reasons. \text{First, the conditioning stream is undifferentiated, leaving no workspace for reasoning.} Existing DiTs concatenate textual and visual tokens into a single joint sequence that is processed uniformly by every transformer block~\citep{peebles2023scalable,esser2024scaling,polyak2024movie}, forcing the same token set to encode both the {high-level editing intent} and the {low-level visual targets} that specify which regions, textures, and motions must change or be preserved~\citep{hertz2022prompt,patashnik2023localizing}.
\text{Second, the reasoning pathway is unsupervised.} Flow-matching and denoising objectives~\citep{ho2020denoising,lipman2022flow} supervise only the predicted velocity field, so the cross-attention maps, which constitute the model's reasoning trace, encoding which text token reads which visual region at which depth~\citep{hertz2022prompt,tang2023daam,chefer2023attend}, are shaped purely by gradients back-propagated through a pixel-level reconstruction loss~\citep{chefer2023attend,rassin2023linguistic,agarwal2023star}.
Thus, the direct symptoms of an attention map are too blunt to resolve the fine-grained correspondences an instruction-faithful edit requires. In other words, the reasoning happens, but it is never taught.

Motivated by this bottleneck, we propose \textbf{\method} to constrain both \emph{what the backbone observes} and \emph{what its internal computations are encouraged to resemble}.
To structure the conditioning stream, we first derive a compact set of \emph{learnable editing tokens} from a MLLM and use them as an abstract carrier of the editing intent. Rather than exposing every block to the same mixture of visual and textual tokens, we route information by depth: shallow blocks attend only to the edit-level tokens, while deeper blocks also receive the native visual and textual tokens for local grounding. This depth-aware routing creates a coarse-to-fine editing process inside the DiT backbone without requiring an explicit symbolic reasoning trace; we refer to it as \emph{Granularity-Routed Token Conditioning} (\S~\ref{sec:grtc}).
To supervise the resulting internal reasoning trace, we further add a parameter-sharing reference branch during training, conditioned on the ground-truth edited video and caption. By maximizing mutual information between the cross-attention maps of the editing branch and this informed reference branch at a randomly sampled depth, the model is encouraged to learn attention patterns that better reflect a successful edit. The reference branch is removed at inference, so \method adds no extra deployment cost; we call this objective \emph{Reference-Anchored Attention Alignment} (\S~\ref{sec:raaa}).

In summary, this paper makes the following contributions:
\begin{itemize}
    \item We present \method, a DiT-native framework for instruction-based video editing that formulates the task as a latent reasoning problem, identifying two structural bottlenecks in existing DiT pipelines that hinder implicit reasoning: undifferentiated conditioning across depths and unsupervised internal reasoning traces.

   \item We propose two complementary components to endow the DiT backbone with implicit reasoning capabilities. \emph{Granularity-Routed Token Conditioning} establishes a coarse-to-fine editing by routing MLLM-derived learnable editing tokens and native visual/textual tokens to different depths, while \emph{Reference-Anchored Attention Alignment} actively supervises this reasoning pathway by maximizing the mutual information between the cross-attention maps of the editing branch and an informed reference branch.

   \item Extensive experiments demonstrate that \method consistently outperforms strong open-source baselines, successfully internalizing the reasoning process to achieve significant gains on localized, compositional, and fidelity-sensitive edits, with ablations confirming the complementarity of the two proposed components.
\end{itemize}
\section{Related Work}

\subsection{Instruction-based Video Editing}
Instruction-based editing has emerged as a natural interface for generative models, allowing users to modify visual content through free-form natural-language commands rather than carefully engineered source--target prompt pairs. Pioneered in the image domain by InstructPix2Pix~\cite{brooks2023instructpix2pix} and its successors~\cite{shagidanov2024grounded,chen2025instruct}, this paradigm has been progressively transferred to video. Early attempts such as InsV2V~\cite{cheng2023consistent} and InstructVid2Vid~\cite{qin2024instructvid2vid} inflate image-level instruction-tuned backbones along the temporal axis, but inherit their limited motion fidelity and suffer from flickering artifacts. More recent efforts push scale along both data and architecture: EffiVED~\cite{zhang2024effived} constructs large instruction-video triplet corpora through automated pipelines, while Se\~norita-2M~\cite{zi2025se} and related million-scale benchmarks expose the task to diverse real-world edits. Despite this progress, these systems are predominantly built upon UNet-based video diffusion priors~\cite{tu2025motionfollower,zhu2025fade} and remain agnostic to the structural properties of modern Diffusion Transformers (DiTs)~\cite{peebles2023scalable}, leaving open the question of how to faithfully ground a textual instruction in the token-level computation of a DiT backbone. More recently, DiT-based methods have begun to emerge~\cite{baiditto,team2025lucy,tan2025omni,he2025openve,lin2026kiwi}, for instance, Ditto~\cite{baiditto} leverages edited keyframes and depth maps, and Kiwi-Edit~\cite{lin2026kiwi} and OpenVE-Edit~\cite{he2025openve} integrate vision-language models to enhance instruction following. 
Despite this progress, all of these methods treat the DiT backbone as a conditioned generator, leaving two structural bottlenecks unaddressed: undifferentiated token conditioning across depths and unsupervised attention geometry, precisely the gaps our method targets.

\subsection{Attention and Feature Alignment in Diffusion Models}
Attention maps and intermediate features inside diffusion backbones have been repeatedly identified as the locus where semantic correspondence between prompt tokens and visual regions emerges, and aligning or manipulating them has become a principled lever for controllable synthesis. Prompt-to-Prompt~\cite{hertz2022prompt} first demonstrated that swapping or reweighting cross-attention maps enables localized image edits without retraining, while Plug-and-Play~\cite{tumanyan2023plug} and MasaCtrl~\cite{cao2023masactrl} generalize this insight by transferring self-attention features and queries across generation passes to preserve structural fidelity. In the video regime, FateZero~\cite{qi2023fatezero}, Video-P2P~\cite{liu2024video}, and TokenFlow~\cite{geyer2023tokenflow} propagate attention and feature correspondences along the temporal axis to suppress flickering and enforce inter-frame consistency, whereas more recent DiT-oriented studies~\cite{mao2026ldt,chen2024gentron} analyze the emergent attention geometry of transformer-based video priors. 

Our approach aligns with this broader programme but fundamentally advances it to benefit video editing tasks by internalizing the reasoning process directly into the DiT backbone, overcoming the brittleness of post-hoc interventions on frozen models. Specifically, we resolve the undifferentiated conditioning bottleneck of standard architectures by (i)~routing instruction tokens to diffusion blocks at a \emph{granularity commensurate with their semantic scope}, (ii)~aligning the cross-branch attention distribution between the reference and the edited stream via the mutual-information objective. Together, these mechanisms endow the model with implicit reasoning capabilities, yielding structurally faithful edits that seamlessly balance instruction compliance with temporal coherence.
\section{Problem Statement}~\label{sec:problem}
We formalize instruction-based video editing as a conditional generation problem that is jointly constrained by a natural-language instruction, the content of a source video, and a set of editing-specific invariances. 
The goal in this section is to make precise what a system is expected to take as input, what it needs to produce.
Let $\mathcal{V} = \{v \mid v \in \mathbb{R}^{T \times H \times W \times 3}\}$ denote videos with $T$ frames of spatial resolution $H \times W$, and let $\mathcal{I}$ denote the natural-language instructions. 

A system for instruction-based video editing is a mapping
\begin{equation}
\small
    f_\theta : \mathcal{V} \times \mathcal{I} \rightarrow \mathcal{\hat{V}}, \qquad \hat{v} = f_\theta(v, \ell),
\end{equation}
parameterized by $\theta$, that takes a source video $v \in \mathcal{V}$ and an instruction $\ell \in \mathcal{I}$ and returns an edited video $\hat{v} \in \mathcal{\hat{V}}$ of the same length and resolution as $v$. 

\paragraph{Problem formulation.} Unlike unconstrained video generation, instruction-based video editing requires the output $\hat{v}$ to be \emph{jointly} faithful to the instruction $\ell$ and to the source video $v$: the edit must realize the semantics of $\ell$ while leaving untouched content intact and staying coherent across time. 
To formalize this objective, we encapsulate these requirements into a unified editing criterion $\mathcal{J}(\hat{v}, v, \ell) \in \mathbb{R}_{\geq 0}$, which is minimized when $\hat{v}$ is simultaneously instruction-faithful, content-preserving, and temporally consistent. The learning process thus entails optimizing the model parameters $\theta$ to minimize this criterion in expectation over the data distribution $\mathcal{D}$:
\begin{equation}
    \min_{\theta}\;\; \mathbb{E}_{(v,\ell) \sim \mathcal{D}} \Bigl[\,\mathcal{J}\bigl(f_\theta(v,\ell),\, v,\, \ell\bigr)\,\Bigr].
    \label{eq:problem}
\end{equation}
Unlike image editing or generic video generation, the constraints in Eq.~\eqref{eq:problem} are inherently coupled: excessive modification to satisfy $\ell$ compromises source fidelity, while independent frame-wise processing degrades temporal coherence. Consequently, naive integration of image editors with video backbones fails to achieve an optimal trade-off, motivating the unified framework proposed in \S\ref{sec:method}.
\begin{figure}[t]
    \centering
    \includegraphics[width=\linewidth]{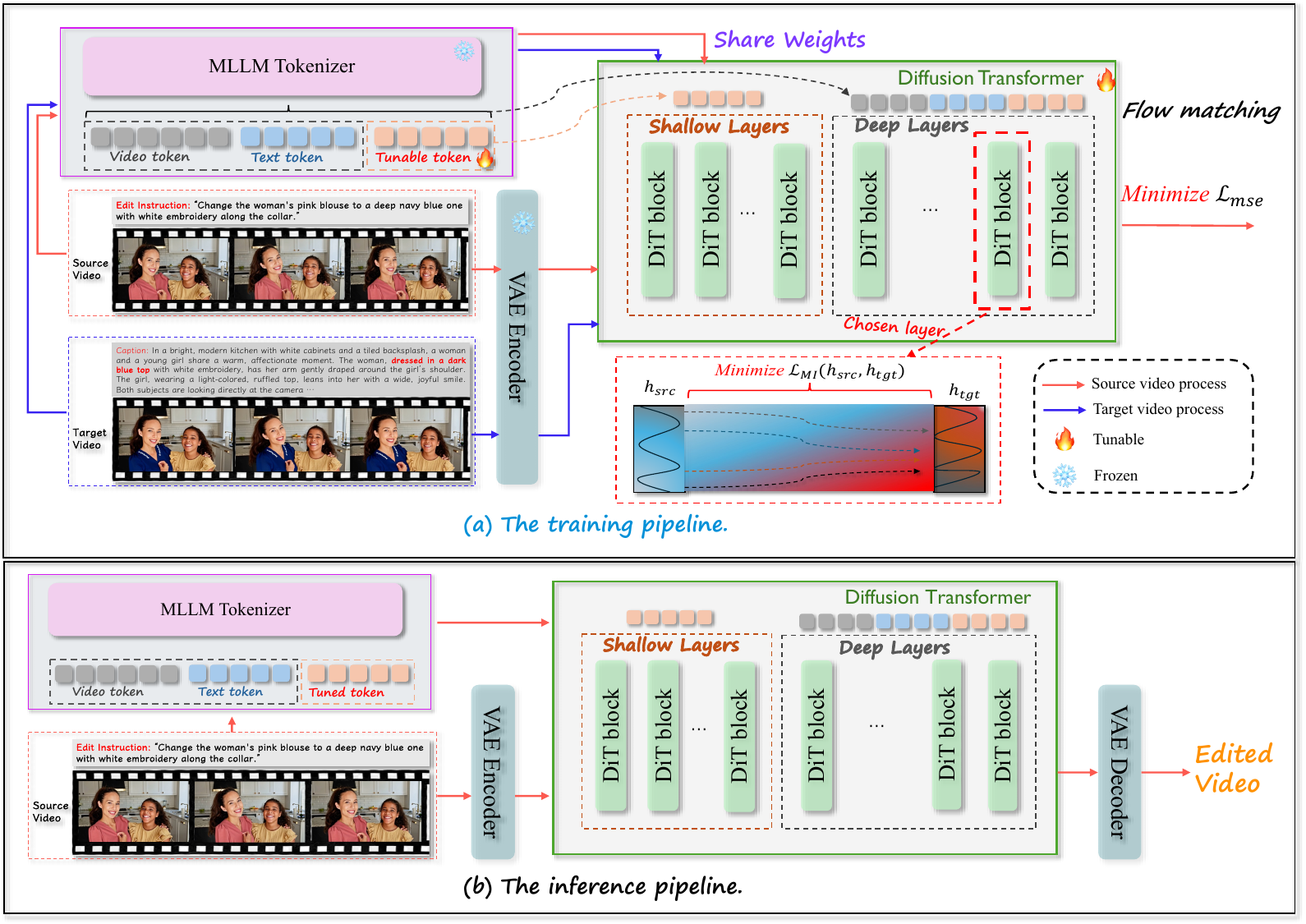}
    \caption{Overview of the \method framework. \textbf{(a) Training:} Given a source video and instruction, we extract learnable editing tokens via an MLLM. The \emph{Granularity-Routed Token Conditioning} module (§\ref{sec:grtc}) routes these abstract tokens to shallow DiT layers to establish global intent, while routing native visual/textual tokens to deeper layers for local refinement. Concurrently, \emph{Reference-Anchored Attention Alignment} (§\ref{sec:raaa}) regularizes the model's internal reasoning by maximizing mutual information between the cross-attention maps of the editing branch and a reference branch conditioned on the target video. \textbf{(b) Inference:} The reference branch is discarded, allowing the model to perform high-fidelity, temporally coherent editing with no additional computational overhead.}
    \label{fig:framework}
    \vspace{-16pt}
\end{figure}
\section{Method}~\label{sec:method}

\vspace{-28pt}
\subsection{Overview of \method}
Figure~\ref{fig:framework} illustrates our framework, a diffusion-transformer realization of the mapping $f_\theta$ introduced in \S\ref{sec:problem}, built around two editing-specific components that directly target the coupled demands of Eq.~\eqref{eq:problem}: a \emph{granularity-routed token conditioning} module (\S\ref{sec:grtc}) that disentangles editing intent across complementary granularities, and a \emph{reference-anchored attention alignment} objective (\S\ref{sec:raaa}) that enforces fine-grained correspondence between the edited and the target video. During training (Figure~\ref{fig:framework}(a)), we tokenize a source video $v$, its instruction $\ell$, and a reference target video $v^\star$ with caption $\ell^\star$ into visual and textual tokens, prepend a small set of \emph{learnable tokens} that act as coarse-grained carriers of editing intent, and route them through the granularity-routed token conditioning module so that the learnable tokens steer global semantics in the shallow transformer layers while the visual and textual tokens refine local appearance in the deeper layers; on top of the standard diffusion denoising loss, the reference-anchored attention alignment term further tightens the model by matching the attention distributions induced by $(v,\ell)$ to those induced by $(v^\star,\ell^\star)$, aligning \emph{where} and \emph{how} the model attends rather than merely \emph{what} it generates. 

At inference (Figure~\ref{fig:framework}(b)), only $(v,\ell)$ are available: they are tokenized, augmented with learnable tokens, and passed through the exact same granularity-routed conditioning, after which the diffusion transformer iteratively denoises a video latent and emits the edited video $\hat{v}$, the reference-anchored alignment acts purely as a training-time regularizer and adds no inference cost, so the framework is as lightweight at deployment as a vanilla diffusion transformer while being substantially more faithful and temporally coherent.

\subsection{Granularity-Routed Token Conditioning} \label{sec:grtc}

As formalized in \S\ref{sec:problem}, instruction-based video editing constitutes a {latent reasoning} problem, wherein the network must jointly infer \emph{which} regions to modify, \emph{what} transformation to apply, and \emph{what} content to preserve within a single generative pass. However, standard text-to-video backbones~\cite{zhang2023adding} inject all conditioning signals uniformly across layers, thereby entangling the \emph{high-level editing intent} with the {low-level visual targets} within a single token set. Consequently, shallow blocks are diverted by pixel-level tokens rather than consolidating global intent, whereas deep blocks are required to reconcile intent and local evidence within a single attention map, which collapses the reasoning pathway and aggravates the fidelity--coherence trade-off in Eq.~\eqref{eq:problem}. Addressing this bottleneck therefore requires an \emph{explicit, pixel-decoupled carrier} of editing intent, together with a \emph{depth-aware delivery} mechanism that routes each granularity to the layers at which it is most informative, thus establishing a coarse-to-fine reasoning chain along the network's depth.
We therefore make that intent explicit by querying an MLLM (see \S\ref{sec:exp}) with the pair $(v,\ell)$ and, following the token-extraction scheme of Kiwi-Edit~\cite{lin2026kiwi}, taking its last-position hidden states as the \emph{learnable editing tokens} $\mathbf{X}_e \in \mathbb{R}^{K \times d}$. Because these positions attend causally over both the visual frames and the instruction inside the MLLM, $\mathbf{X}_e$ aggregates a globally-contextualized, vision-grounded summary of the intended edit, a far more abstract carrier of intent than the raw textual tokens. Concatenating the three streams yields the full conditioning sequence $\mathbf{X} = [\mathbf{X}_e;\, \mathbf{X}_v;\, \mathbf{X}_\ell]$, in which high- and low-level information are now \emph{separately addressable}.


Rather than treating $\mathbf{X}_e$, $\mathbf{X}_v$, and $\mathbf{X}_\ell$ symmetrically at every depth, we exploit the well-documented observation~\cite{ho2024block} that shallow transformer blocks capture coarse, globally-shared semantics while deeper blocks specialize in fine-grained spatial and appearance details. Concretely, let the diffusion transformer consist of $L$ blocks and let $L_s \in \{1,\dots,L\}$ be a routing depth. For the shallow stage $\ell_{\mathrm{idx}} \leq L_s$, we expose \emph{only} the learnable editing tokens $\mathbf{X}_e$ to the cross-attention of each block, so that the global editing intent unambiguously shapes the early denoising trajectory without being diluted by thousands of pixel-level tokens. For the deep stage $\ell_{\mathrm{idx}} > L_s$, we expand the condition to the \emph{full} sequence $[\mathbf{X}_e;\,\mathbf{X}_v;\,\mathbf{X}_\ell]$, so that the intent carried by $\mathbf{X}_e$ remains accessible as a persistent anchor while the native visual and textual tokens ground it onto concrete regions, textures, and motions. Formally, the cross-attention context at block $\ell_{\mathrm{idx}}$ is
\begin{equation}
\small
    \mathbf{C}^{(\ell_{\mathrm{idx}})} \;=\;
    \begin{cases}
        \mathbf{X}_e, & \ell_{\mathrm{idx}} \leq L_s, \\[2pt]
        [\,\mathbf{X}_e;\,\mathbf{X}_v;\,\mathbf{X}_\ell\,], & \ell_{\mathrm{idx}} > L_s.
    \end{cases}
    \label{eq:routing}
\end{equation}
This depth-wise split is a progressive disclosure rather than a partition: shallow layers commit to a global editing intent from $\mathbf{X}_e$ alone, while deeper layers refine region-level appearance by re-admitting the native tokens \emph{on top of} this anchor.
No information in $\mathbf{X}$ is ever discarded and each group participates exactly where it is most informative. As a result, shallow layers are freed from having to distill intent out of long, noisy token sequences, while deep layers can refine region-level appearance against an anchor $\mathbf{X}_e$ that has already been committed to, which directly relaxes the coupling between instruction faithfulness and source preservation identified in \S\ref{sec:problem}.

\subsection{Reference-Anchored Attention Alignment}
\label{sec:raaa}
Granularity-routed conditioning controls \emph{what} each depth observes, but it does not specify \emph{where} the model should attend. This leaves the DiT's cross-attention maps, its implicit reasoning trace~\cite{hertz2022prompt,tang2023daam}, supervised only through the flow-matching loss~\cite{lipman2022flow}. Yet the standard flow-matching objective~\cite{lipman2022flow} supervises only the denoised velocity field, leaving this trace shaped purely by gradients back-propagated through a pixel-level reconstruction loss. The network is therefore free to adopt any attention geometry that happens to reduce that loss, and empirically these unconstrained geometries collapse onto coarse token-to-region associations that cannot resolve the fine-grained correspondences the instruction requires: \emph{which} regions must change, \emph{which} must be preserved, and \emph{how} the change must persist across frames. In other words, \textit{the reasoning happens but is never taught.}
We therefore introduce an attention-level training signal built from two parts: (i) a reference anchor that represents a successful reasoning trace, and (ii) a correspondence objective that aligns the editing branch to this anchor without forcing exact imitation.
We obtain the first ingredient by exploiting a resource that is already available at training time: for each pair $(v,\ell)$, the dataset provides a \emph{reference target} $(v^\star, \ell^\star)$, where $v^\star$ is a ground-truth edited video and $\ell^\star$ its caption. Intuitively, $(v^\star,\ell^\star)$ describes the \emph{outcome} of a successful reasoning chain while $(v,\ell)$ describes its \emph{premise}, so a well-trained editor should internally transport the latter toward the former and this transport should be visible in the attention trace itself. We run a second forward pass through the \emph{same} diffusion transformer, with shared parameters $\theta$, conditioned on $(v^\star,\ell^\star)$ using the tokenization pipeline of \S\ref{sec:grtc}. The resulting cross-attention maps form an \emph{in-distribution} reasoning trace for a solved edit, providing a natural anchor for the editing branch. This branch is used only to extract training-time traces and is discarded at inference, adding no parameters or deployment cost.


The second ingredient is the correspondence objective that couples the two reasoning traces. Let $\mathbf{A}^{(\ell_{\mathrm{idx}})}(v,\ell) \in \mathbb{R}^{S \times d}$ and $\mathbf{A}^{(\ell_{\mathrm{idx}})}(v^\star,\ell^\star) \in \mathbb{R}^{S \times d}$ denote the cross-attention output features produced at block $\ell_{\mathrm{idx}} \in \{1,\dots,L\}$ by the editing branch and the reference branch, respectively, with each of the $S$ spatio-temporal positions treated as a sample of the trace at that depth.To avoid the editing trace imitating the reference distribution element by element and leaks nuisance idiosyncrasies that have nothing to do with the reasoning itself, we adopt a \emph{mutual-information} criterion, which depends only on the \emph{shared} information between the two traces and is symmetric in its two arguments, yielding a supervision signal that is strong on the reasoning structure: 
\begin{equation}
\small
    \mathcal{L}_{\mathrm{align}}
    \;=\;
    -\,\mathbb{E}_{\ell_{\mathrm{idx}} \sim \mathcal{U}\{1,\dots,L\}}
    \Bigl[\,
        \hat{I}\!\left(
            \mathbf{A}^{(\ell_{\mathrm{idx}})}(v,\ell)\,;\,
            \mathbf{A}^{(\ell_{\mathrm{idx}})}(v^\star,\ell^\star)
        \right)
    \,\Bigr],
    \label{eq:align}
\end{equation}
where $\hat{I}(\cdot\,;\cdot)$ is a differentiable lower bound on the mutual information $I(\cdot\,;\cdot)$ between its two arguments. We instantiate it with a \emph{symmetric InfoNCE}~\cite{oord2018representation} estimator on $\ell_2$-normalized token features: writing $\tilde{\mathbf{a}}_i$ and $\tilde{\mathbf{a}}^\star_i$ for the normalized $i$-th rows of $\mathbf{A}^{(\ell_{\mathrm{idx}})}(v,\ell)$ and $\mathbf{A}^{(\ell_{\mathrm{idx}})}(v^\star,\ell^\star)$, position $i$ in the editing branch is paired with position $i$ in the reference branch as the positive match, while the remaining $S-1$ positions act as negatives, yielding
\begin{equation}
\small
    \hat{I}\bigl(\mathbf{A};\mathbf{A}^\star\bigr)
    \;=\;
    \frac{1}{2S}\sum_{i=1}^{S}
    \Biggl[
        \log\!\frac{\exp\!\bigl(\tilde{\mathbf{a}}_i^{\!\top}\tilde{\mathbf{a}}^\star_i/\tau\bigr)}
                   {\sum_{j=1}^{S}\exp\!\bigl(\tilde{\mathbf{a}}_i^{\!\top}\tilde{\mathbf{a}}^\star_j/\tau\bigr)}
        \;+\;
        \log\!\frac{\exp\!\bigl(\tilde{\mathbf{a}}_i^{\!\top}\tilde{\mathbf{a}}^\star_i/\tau\bigr)}
                   {\sum_{j=1}^{S}\exp\!\bigl(\tilde{\mathbf{a}}_j^{\!\top}\tilde{\mathbf{a}}^\star_i/\tau\bigr)}
    \Biggr],
    \label{eq:infonce}
\end{equation}
where $\tau$ is a temperature hyper-parameter. The gradients flow \emph{only} through the editing branch, the reference branch is detached so that its trace behave as a fixed anchor. Intuitively, Eq.~\eqref{eq:infonce} pulls each editing-branch token toward the reference-branch token at the \emph{same} spatial--temporal position while pushing it away from all other positions,  which is precisely the positional correspondence a faithful reasoning chain must establish in Eq.~\eqref{eq:problem}. 
We sample a single depth $\ell_{\mathrm{idx}}$ uniformly at each iteration, yielding an unbiased stochastic estimator of the full layer-wise mutual information whose variance acts as an implicit regularizer akin to stochastic depth.

\paragraph{Training objective.}
The final objective couples the standard flow-matching loss $\mathcal{L}_{\mathrm{fm}}$ with the proposed alignment term,
\begin{equation}
    \mathcal{L}
    \;=\;
    \mathcal{L}_{\mathrm{fm}}
    \;+\;
    \lambda_{\mathrm{align}}\,\mathcal{L}_{\mathrm{align}},
    \label{eq:total}
\end{equation}
where $\lambda_{\mathrm{align}} \geq 0$ trades off pixel-level fidelity against reasoning-level correspondence. Whereas $\mathcal{L}_{\mathrm{fm}}$ supervises only the network's output, $\mathcal{L}_{\mathrm{align}}$ supervises its implicit reasoning trace itself, teaching the editing branch \emph{where} to attend by maximizing the information it shares with a trace that is known to solve the edit, rather than letting the trace silently settle into whatever pattern minimizes reconstruction. Together with the granularity routing of \S\ref{sec:grtc}, which provides the substrate on which such a coarse-to-fine chain can be staged, this converts the reference target from a passive reconstruction goal into an active pedagogical signal for the reasoning chain, and closes the loop on the coupled faithfulness--preservation--coherence demands of Eq.~\eqref{eq:problem}.
\section{Experiment} \label{sec:exp}

We summarize the key protocol below and defer the full configuration to Appendix~\ref{app:exp_settings}.
\paragraph{Data Construction.}
We build our training corpus from three public video-editing datasets: OpenVE-3M~\cite{he2025openve}, ReCo~\cite{zhang2025region}, and Ditto-1M~\cite{bai2025scaling}, which together cover appearance modification, object insertion/removal, stylization, and motion-preserving attribute change. To balance editing categories and prevent any single source from dominating, we draw a stratified sample of $10{,}000$ clips per category from each dataset, yielding $120{,}000$ instruction--video--target triplets. Since the released captions are heterogeneous, we re-annotate every target video with Qwen3-VL-8B-Instruct~\cite{bai2025qwen3} to obtain unified, visually grounded descriptions compatible with the reference branch in \S\ref{sec:raaa}.

\paragraph{Training Settings}
We instantiate \method with Wan2.2-TI2V-5B~\cite{wan2025wan} ($L{=}34$ DiT blocks) as the video backbone and Qwen2.5-VL-3B~\cite{bai2025qwen3} as the MLLM that yields $K{=}512$ learnable editing tokens, and train on $81$-frame clips at $720{\times}1280$. The routing split is $L_s{=}17$ and the alignment term uses $\lambda_{\mathrm{align}}{=}0.75$ with InfoNCE temperature $\tau{=}0.07$. The DiT and the MLLM-to-DiT connectors are fully fine-tuned; the MLLM language tower is adapted with rank-$64$ LoRA~\cite{hu2022lora} and its vision encoder is frozen. We optimize Eq.~\eqref{eq:total} with AdamW at learning rate $1{\times}10^{-5}$ for one epoch in \texttt{bf16} on $8$ NVIDIA H800 GPUs.

\paragraph{Evaluation.}
We evaluate \method on \textbf{OpenVE-Bench} against representative open-source video editors: VACE~\cite{jiang2025vace}, OmniVideo~\cite{yang2026omni}, ICVE~\cite{liao2025context}, Lucy-Edit~\cite{team2025lucy}, DITTO~\cite{baiditto}, and Kiwi-Edit~\cite{lin2026kiwi}. Since surrogate metrics correlate weakly with human judgment on open-ended edits, we adopt an \emph{MLLM-as-a-judge} protocol with two complementary judges, Gemini-3.1-Pro~\cite{comanici2025gemini} and Qwen3.5-VL-35B~\cite{bai2025qwen3}, each scoring every video on a $1$--$100$ scale along three axes reflecting Eq.~\eqref{eq:problem}: \emph{Instruction Compliance} (IC), \emph{Consistency Fidelity} (CF), and \emph{Visual Quality} (VQ). We report per-category scores averaged across the two judges; prompt templates and aggregation details are in Appendix~\ref{app:exp_eval}.
\begin{table}[htbp]
    \centering
    \vspace{-8pt}
    \caption{Quantitative comparison on \textbf{OpenVE-Bench}. All scores are on a $1$-to-$100$ Likert scale and are averaged across the two MLLM judges (Qwen3.5-VL-35B and Gemini-3.1-Pro). Among open-source methods, \textbf{bold} marks the best and \underline{underline} marks the second-best.}
    \label{tab:quantitative_results}
    \resizebox{\linewidth}{!}{%
    \begin{tabular}{l c c c | c c c |c c c |c cc|ccc}
        \toprule
        \multirow{2}{*}{Method} & \multirow{2}{*}{\#Params.} & \multirow{2}{*}{\#Reso.} & \multirow{2}{*}{Overall}  & \multicolumn{3}{c|}{\textbf{\makecell{Local\\Change}}} & \multicolumn{3}{c|}{\textbf{\makecell{Local\\Remove}}} & \multicolumn{3}{c|}{\textbf{\makecell{Local\\Add}}} & \multicolumn{3}{c}{\textbf{\makecell{Global\\Style}}}\\
    
        ~ &~ & ~& ~& IC & CF & VQ & IC & CF & VQ & IC & CF & VQ & IC & CF & VQ \\
        \midrule
        \multicolumn{16}{c}{\textbf{Gemini-3.1-Pro}} \\
        \midrule
        VACE~\cite{jiang2025vace}       & 14B  & $1280{\times}720$ & 16.66 & 37.15 & 6.46 & 18.92 &  5.76 & 5.37 & 9.75 & 11.34 & 5.22 & 14.78 & 29.79 & 15.10 & 40.22\\
        Lucy-Edit~\cite{team2025lucy}  & 5B   & $1280{\times}704$ & 49.24 & 68.09 & 57.14 & 63.37 & 7.68 & 44.69 & 46.80 &37.64 & 38.30 & 39.16 &16.17 & 84.38 & 87.48 \\
        ICVE~\cite{liao2025context}       & 13B  & $384{\times}240$  & 52.36 & 61.58 & 42.85 & 64.26&  38.73 & 43.05 & 58.68 & 38.13 & 32.54 & 51.04& 60.66 & 58.93 & 77.91\\
        DITTO~\cite{baiditto}      & 14B  & $832{\times}480$  & 36.82 & 48.29 & 16.52 & 57.43& 6.05 & 11.81 & 36.86 & 44.25 & 8.36 &49.70&  53.79 & 37.00 & 71.76\\
        Kiwi-Edit~\cite{lin2026kiwi} & 5B  & $1280{\times}704$  & \underline{61.35} & 81.38 & 62.23 & 71.97 & 66.63 & 48.27 & 46.69 &47.96 & 40.46 & 33.45 &  70.86 & 84.19 & 82.07\\
           \midrule
        \textbf{\method (Ours)} & 5B & $1280{\times}704$ & \textbf{62.48} &  69.77 & 70.35 & 69.42 & 71.71 & 41.59 & 42.80 & 49.46 & 55.58 & 45.09 & 67.78 & 82.76 & 83.48\\

        \midrule
        \multicolumn{16}{c}{\textbf{Qwen3.5-VL-35B}} \\
        \midrule
        VACE~\cite{jiang2025vace}       & 14B  & $1280{\times}720$ & 28.41 & 32.65 & 16.77 & 61.28 & 8.15 &  8.15 & 14.42 & 14.24 & 10.81 & 34.42 &  36.64 & 36.43 & 66.91\\
        Lucy-Edit~\cite{team2025lucy}  & 5B   & $1280{\times}704$ & 49.24 & 68.09 & 57.14 & 63.37 & 7.68 & 44.69 & 46.80 &37.64 & 38.30 & 39.16 &16.17 & 84.38 & 87.48 \\
        ICVE~\cite{liao2025context}       & 13B  & $384{\times}240$  & 66.56 & 68.46 & 76.35 & 86.11 & 52.37 & 63.61 & 69.05 & 52.34 & 59.22 & 73.69 & 60.66 & 58.93 & 77.91 \\
        DITTO~\cite{baiditto}      & 14B  & $832{\times}480$  & 51.95 & 51.15 & 41.63 & 78.49 &  11.20 & 25.78 &  37.95 & 45.67 & 35.70 & 62.30 & 74.36 & 73.64 & 85.57\\
  
        Kiwi-Edit~\cite{lin2026kiwi} & 5B  & $1280{\times}704$ & \underline{75.89} & 85.28 & 85.46 & 88.05 & 82.29 & 63.58 & 64.44 &63.94 & 63.33 & 73.84 & 73.02 & 86.66 & 80.79\\
        \midrule
        \textbf{\method (Ours)} & 5B & $1280{\times}704$ & \textbf{77.42} & 78.71 & 86.43 & 89.92 & 84.32 & 67.08 &  67.93&59.49 & 71.73 & 79.67& 72.50 & 87.64 & 83.64 \\
        \bottomrule
    \end{tabular}%
    }
    \vspace{-8pt}
\end{table}

\subsection{Main Results}

\paragraph{Quantitative Results.}

Table~\ref{tab:quantitative_results} reports per-category results under two complementary MLLM judges. \method achieves the best overall score under both evaluators, reaching 77.42 with Qwen3.5-VL-35B and 62.48 with Gemini-3.1-Pro. Specifically, \method is strongest on the most demanding categories, \emph{Local Add} and \emph{Local Change}, where successful editing requires both precise modification and preservation of surrounding content. This matches the intended coarse-to-fine division of labor: shallow blocks establish edit intent, and deeper blocks refine local appearance and motion. \method also performs strongly on \emph{Global Style}, showing that the same routing strategy supports coherent scene-level transformations. On \emph{Local Remove}, it remains competitive, though with a smaller margin under one judge, likely reflecting a mild source-preserving bias from attention alignment. Overall, while baselines are typically uneven across tasks, \method is consistently strong across task types.

\begin{figure}[htbp]
    \centering
    \includegraphics[width=\linewidth]{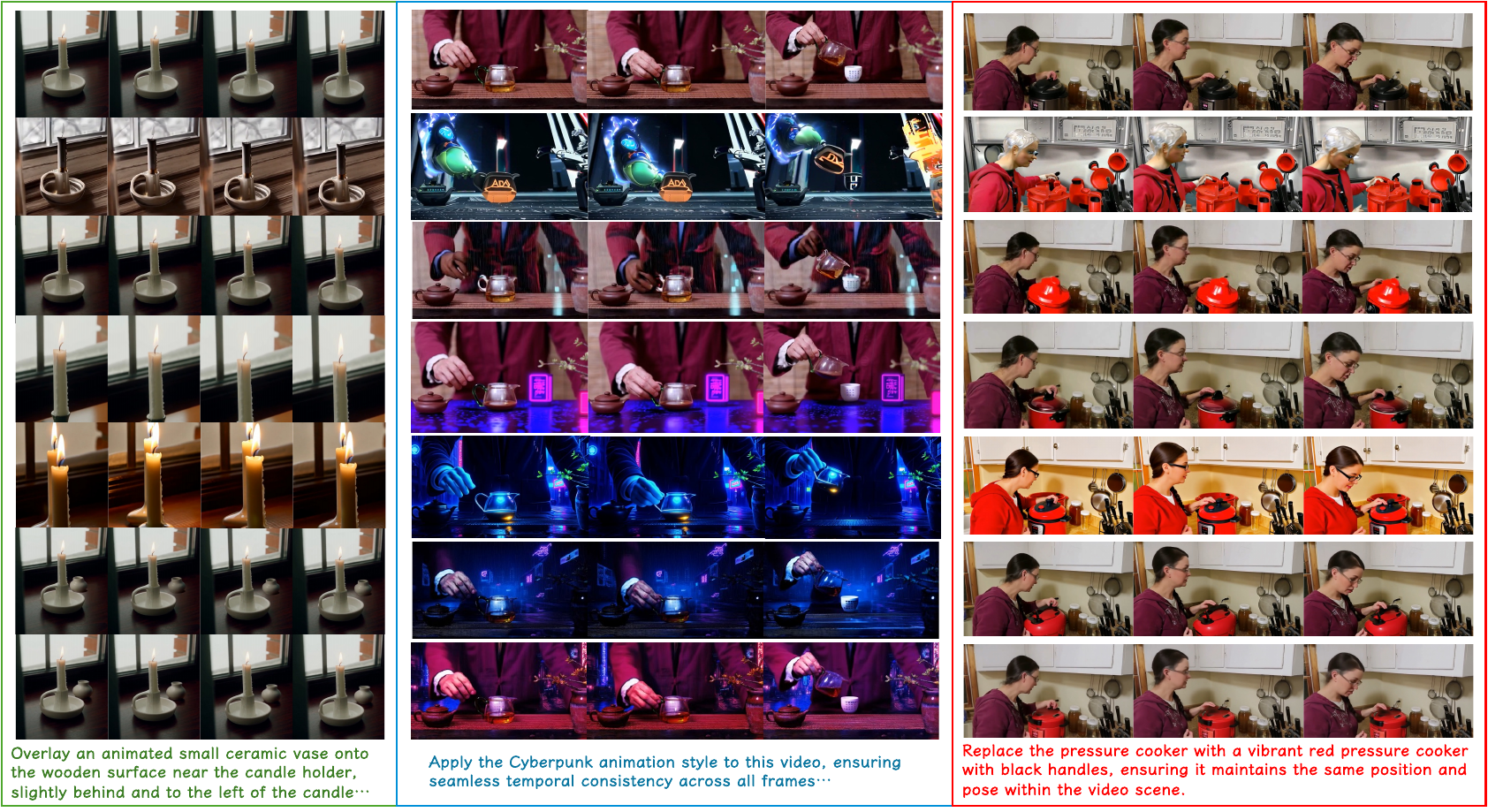}
    \caption{Qualitative comparison on \textbf{OpenVE-Bench} across the four edit categories. \method localizes the edit precisely, preserves non-edited content, and remains temporally stable, whereas baselines display mislocalization, leakage, residual traces, or over-stylization.}
    \label{fig:qualitative}
    \vspace{-12pt}
\end{figure}

\paragraph{Qualitative Results.}
Figure~\ref{fig:qualitative} compares \method against the three most recent baselines, with the full comparison deferred to Appendix~\ref{app:visual}. \method places edits precisely while preserving identity and layout, a direct consequence of \emph{Granularity-Routed Token Conditioning}, which lets shallow blocks lock in editing intent before deeper blocks refine appearance. On \emph{Local Remove} and \emph{Global Style}, it further yields clean erasures and temporally stable restylings, consistent with \emph{Reference-Anchored Attention Alignment} sharpening the backbone's attention geometry. These trends mirror Table~\ref{tab:quantitative_results}, confirming that the gains stem from improved implicit reasoning rather than any MLLM-judge bias.

\subsection{Ablation and Parameter Study} \label{sec:abla}

\begin{figure}[h]
    \centering
    \begin{minipage}[c]{0.38\linewidth}
        \centering
        \captionof{table}{Ablation on \textbf{OpenVE-Bench} under the Qwen3.5-VL-35B. Scores are per-category averages over the three evaluation axes; \emph{Overall} further averages across the four categories.}
        \label{tab:ablation}
        \resizebox{\linewidth}{!}{%
        \begin{tabular}{l c | c c c c}
            \toprule
            Variant & Overall & \makecell{Local\\Change} & \makecell{Local\\Remove} & \makecell{Local\\Add} & \makecell{Global\\Style} \\
            \midrule
            \textbf{\method} & \textbf{77.42} & \text{85.02} & \text{73.11} & \text{70.29} & \text{81.29} \\
            \midrule
            \multicolumn{6}{l}{\emph{(a) Component ablations}} \\
            w/o GRTC\quad($L_s{=}L$)                                  & 74.76 & 84.59 & 71.02 & 65.64 & 77.80 \\
            w/o RAAA\quad($\lambda_{\mathrm{align}}{=}0$)                  & 76.06 & 84.94 & 73.48 & 65.74 & 80.11 \\
            w/o both                             & 71.71 & 83.88 & 56.18 & 67.06 & 79.96 \\
            \midrule
            \multicolumn{6}{l}{\emph{(b) Routing depth $L_s$\; (with $\lambda_{\mathrm{align}}{=}0.75$ fixed)}} \\
            $L_s{=}8$                                                      & 74.13 & 86.48 & 67.95 & 63.50 & 78.61 \\
            \underline{$L_s{=}17$\quad(ours)}                              & \underline{77.42} & \underline{85.02} & \underline{73.11} & \underline{70.29} & \underline{81.29} \\
            $L_s{=}25$                                                     & 69.83 & 80.55 & 63.87 & 58.98 & 76.05 \\
            $L_s{=}L$\quad (w/o GRTC)                                & 74.76 & 84.59 & 71.02 & 65.64 & 77.80 \\
            \midrule
            \multicolumn{6}{l}{\emph{(c) Alignment weight $\lambda_{\mathrm{align}}$\; (with $L_s{=}15$ fixed)}} \\
            $\lambda_{\mathrm{align}}{=}0$\quad(w/o RAAA)                & 76.06 & 84.94 & 73.48 & 65.74 & 80.11 \\
            $\lambda_{\mathrm{align}}{=}0.25$                              & 75.09 & 83.44 & 71.85 & 65.65 & 79.45 \\
            $\lambda_{\mathrm{align}}{=}0.50$                              & 72.75 & 82.94 & 64.67 & 65.27 & 78.13 \\
            \underline{$\lambda_{\mathrm{align}}{=}0.75$\quad(ours)}       & \underline{77.42} & \underline{85.02} & \underline{73.11} & \underline{70.29} & \underline{81.29} \\
            $\lambda_{\mathrm{align}}{=}1.00$                              & 73.37 & 81.76 & 71.18 & 61.61 & 78.93 \\
            \bottomrule
        \end{tabular}%
        }
    \end{minipage}%
    \hfill
    \begin{minipage}[c]{0.6\linewidth}
        \centering
        \includegraphics[width=0.95\linewidth]{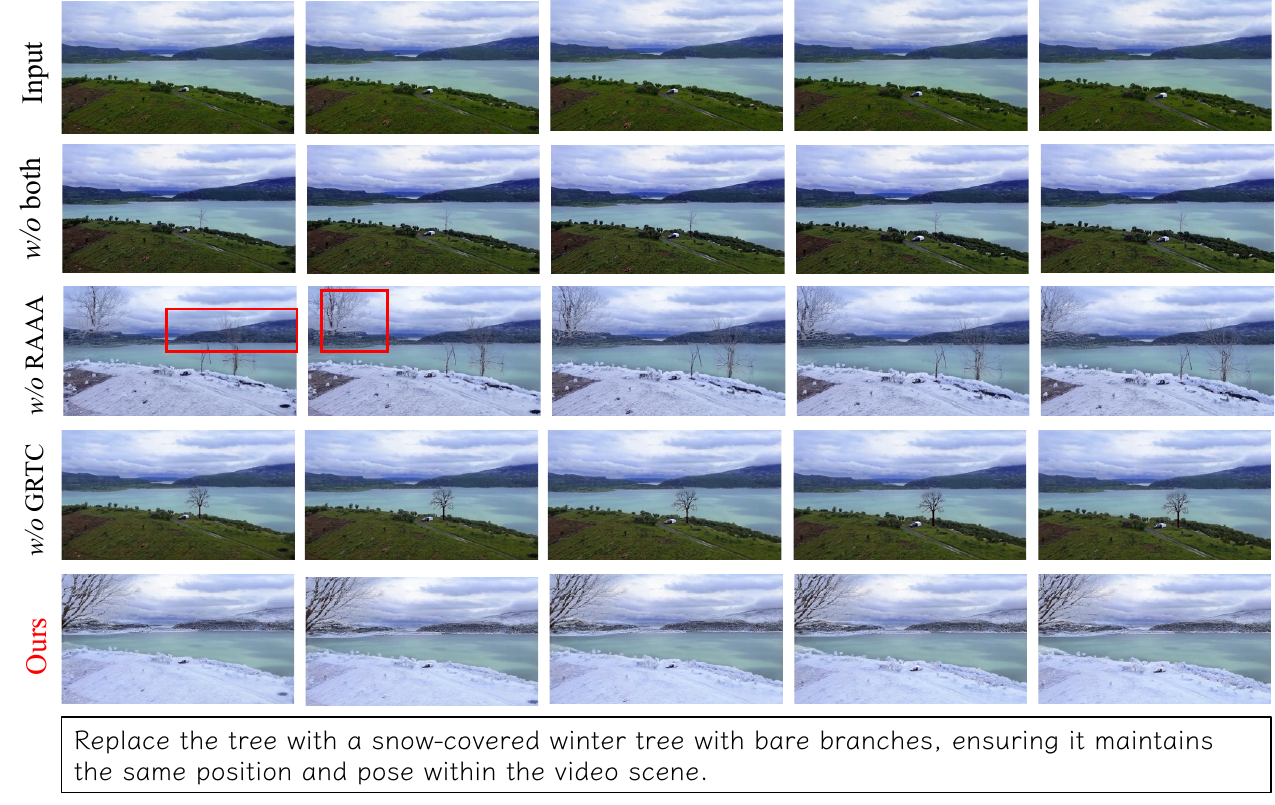}
        \captionof{figure}{Qualitative examples of the component ablations reported in Table~\ref{tab:ablation}.}
        \label{fig:ablation_curve}
    \end{minipage}
    \vspace{-8pt}
\end{figure}

We now isolate the two mechanisms introduced in \S\ref{sec:method} (GRTC refers to \S\ref{sec:grtc}, RAAA refers to \S\ref{sec:raaa}), and examine whether they correct the two structural bottlenecks identified in \S\ref{sec:intro}: an undifferentiated conditioning stream and an unsupervised internal reasoning trace. All variants are trained under the recipe of \S\ref{sec:exp}, evaluated on \text{OpenVE-Bench} under the Qwen3.5-VL-35B judge, and reported as the per-category score averaged over the three evaluation axes; Table~\ref{tab:ablation} groups the results into (a) component ablations, (b) a sweep over the GRTC routing depth $L_s$ of Eq.~\eqref{eq:routing}, and (c) a sweep over the RAAA weight $\lambda_{\mathrm{align}}$ of Eq.~\eqref{eq:total}.

\paragraph{Ablation Analysis} Table~\ref{tab:ablation} and Figure~\ref{fig:ablation_curve} expose distinct failure modes of the two components. Removing GRTC degrades the categories that require a clear global editing intent, notably \emph{Local Add} and \emph{Global Style}, and the resulting edits often drift from the target region or leak style into preserved content. Removing RAAA yields a smaller overall drop, but the degradation is concentrated in categories that depend on fine-grained positional correspondence, consistent with the role of Eq.~\eqref{eq:infonce}. Removing both components leads to a substantially larger failure, with \emph{Local Remove} dropping by nearly seventeen points overall; Figure~\ref{fig:ablation_curve} further shows compounded residual traces and structural corruption, rather than a simple superposition of the two artifacts.

\paragraph{Parameter Analysis} The parameter sweeps support the chosen setting as the best operating point. Performance peaks at an intermediate routing depth $L_s$, while both smaller and larger values reduce effectiveness, consistent with the need to separate early global-intent formation from later local refinement. Performance is also non-monotonic in $\lambda_{\mathrm{align}}$: a moderate alignment weight improves editing quality, whereas excessive weighting biases the model toward source preservation and weakens instruction following, in line with the auxiliary role of $\mathcal{L}_{\mathrm{align}}$ discussed in \S\ref{sec:raaa}. A detailed per-axis breakdown and additional sweep results are provided in Appendix~\ref{app:add_res}.

\subsection{Method Analysis}

We further examine whether the internal behavior of the DiT matches the reasoning mechanism introduced in \S\ref{sec:method}. For GRTC, we measure how cross-attention mass is distributed between the learnable editing tokens $\mathbf{X}_e$ and the native visual/textual tokens across transformer depth, together with the spatial entropy of the resulting attention maps. For RAAA, we compare the attention traces of the editing and reference branches using cosine similarity, and JS divergence, and contrast them with the original base model.
\begin{figure}[h]
    \centering
    \vspace{-8pt}
    \begin{subfigure}{0.52\textwidth}
          \includegraphics[width=\linewidth]{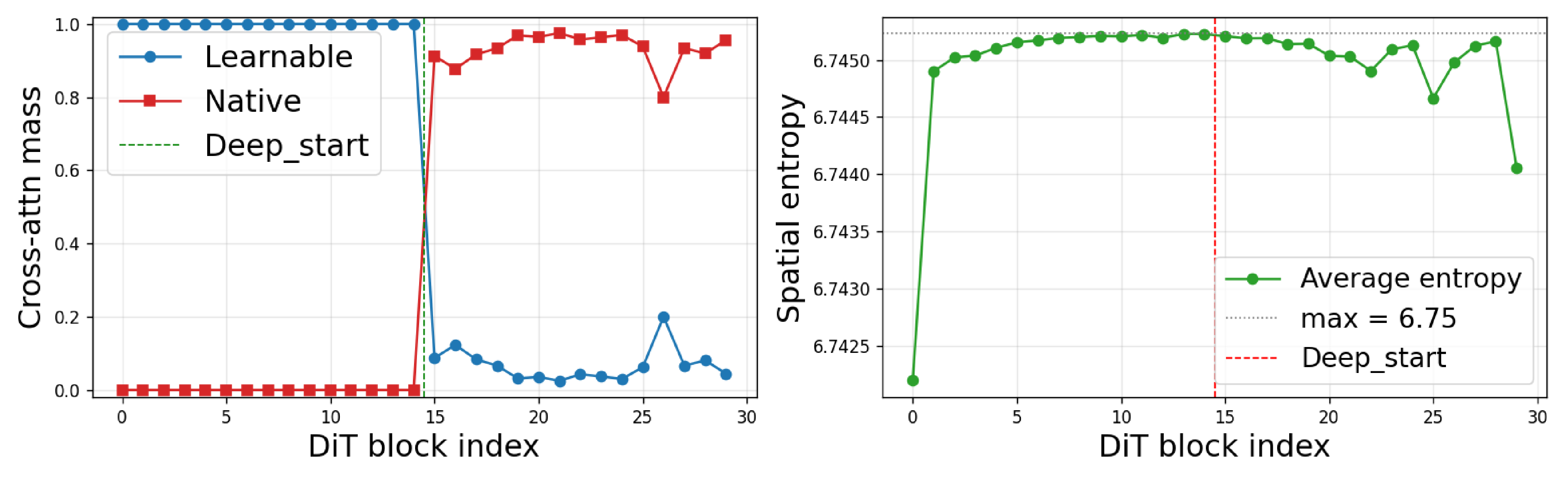}
          \caption{Depth-wise effect of GRTC.}
    \end{subfigure}
    \hfill
    \begin{subfigure}{0.47\textwidth}
          \includegraphics[width=\linewidth]{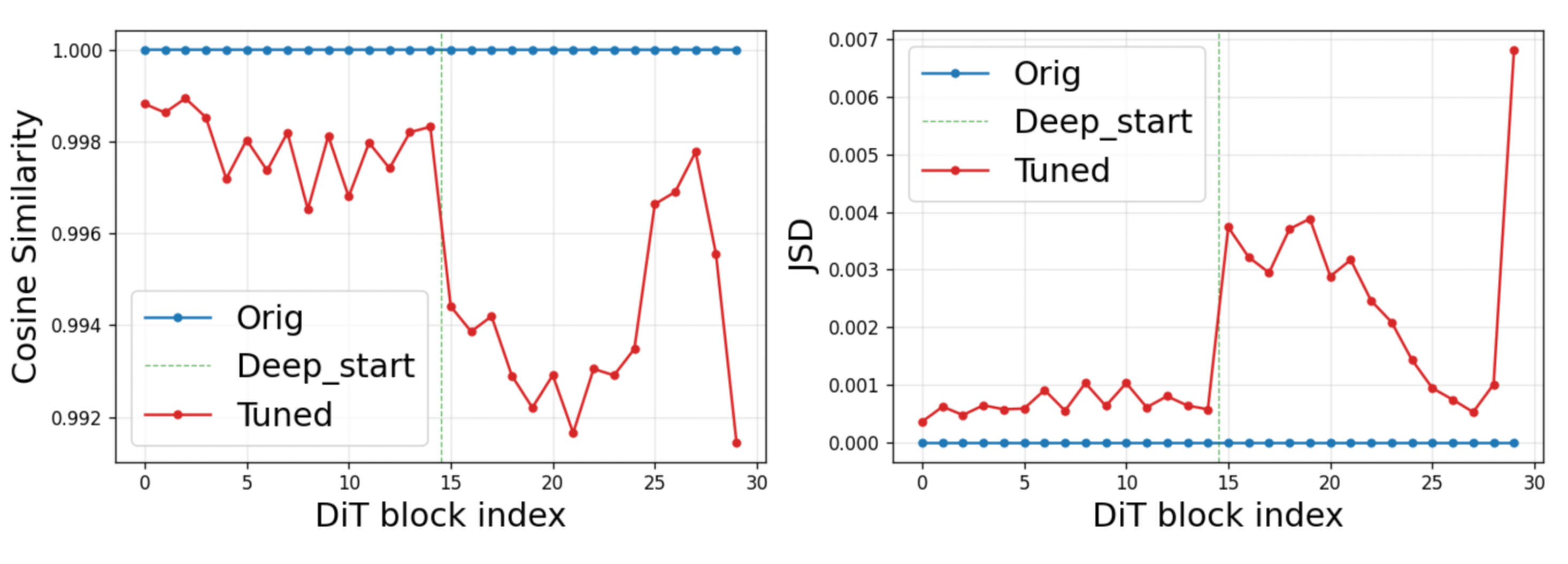}
          \caption{Attention alignment induced by RAAA.}
    \end{subfigure}
    \caption{Mechanistic analysis of \method. (a) Attention mass on learnable tokens and native visual/textual tokens across DiT depth, with spatial attention entropy. (b) Cosine similarity and JS divergence between the editing and reference attention traces, compared with the base model.}
    \label{fig:analysis}
    \vspace{-12pt}
\end{figure}

Figure~\ref{fig:analysis}(a) provides direct evidence for the coarse-to-fine routing in Eq.~\eqref{eq:routing}. Early blocks concentrate attention on the learnable editing tokens $\mathbf{X}_e$, showing that the model first uses the MLLM-derived tokens to establish the global edit intent. After the routing boundary, attention shifts to native visual/textual tokens and becomes spatially sharper, with lower entropy, indicating that this intent is grounded into localized regions, appearances, and motions. Thus, GRTC turns conditioning from a uniform token stream into an explicit depth-wise reasoning path.
RAAA makes this path more reliable by supervising the attention trace itself, as shown in Figure~\ref{fig:analysis}(b). Compared with the base model, the trained model shows stronger agreement between the editing and reference branches, while retaining non-zero divergence; hence, it learns the reference branch's correspondence structure without collapsing into direct imitation. Together with Table~\ref{tab:ablation}, these results support the complementarity of the two components: GRTC provides the reasoning workspace, whereas RAAA stabilizes its attention geometry, leading to balanced gains in instruction following, source preservation, and visual quality.

\section{Conclusion}
This paper studies instruction-based video editing as a latent reasoning problem for Diffusion Transformers. We show that existing DiT editors are limited by two structural bottlenecks: conditioning streams are used uniformly across depth, and the attention traces that mediate editing decisions receive little direct supervision. To address these issues, we propose \textbf{\method}, a DiT-native framework that separates editing intent from fine-grained visual evidence through \emph{Granularity-Routed Token Conditioning}, and regularizes the resulting reasoning trace with \emph{Reference-Anchored Attention Alignment}. The reference branch is used only during training, so the improved supervision introduces no additional inference cost.
Experiments on \text{OpenVE-Bench} demonstrate that \method achieves the strongest overall performance among open-source baselines, with consistent gains on localized, compositional, and fidelity-sensitive edits. Ablation studies and attention analyses further support the design: depth-wise routing provides a structured workspace for coarse-to-fine reasoning, while reference-anchored alignment improves the reliability of the model's internal correspondence. These results suggest that explicitly shaping the internal reasoning process of generative backbones is a promising direction for robust and controllable video editing.


{
\small
\bibliographystyle{abbrv}
\bibliography{reference}
}

%
\newpage
\appendix

\section{Related Work}
\subsection{Reference-guided Video Generation and Editing}
A complementary line of work conditions generation on an auxiliary visual reference, supplying appearance or motion cues that are difficult to specify in language alone. In the image domain, IP-Adapter~\cite{ye2023ip} and subsequent subject-driven variants~\cite{xiao2025fastcomposer} inject reference features through decoupled cross-attention, whereas Q-Former, style resamplers~\cite{swetha2024x} compress reference signals into a compact set of learnable queries for downstream conditioning. Extending these ideas to video, VideoBooth~\cite{jiang2024videobooth}, Still-Moving~\cite{chefer2024still}, and MotionDirector~\cite{zhao2024motiondirector} disentangle subject appearance from motion dynamics, while ConsistI2V~\cite{ren2024consisti2v} and Phantom~\cite{liu2025phantom} exploit a reference frame or clip to enforce temporal coherence. Training-free alternatives such as AnyV2V~\cite{ku2024anyv2v} and related first-frame-anchored pipelines~\cite{ouyang2024i2vedit} propagate edits from a single edited keyframe across the clip via feature injection. These methods convincingly demonstrate that a reference branch can stabilize identity and layout, yet they typically treat the reference as a \emph{global} conditioning signal and do not explicitly align its spatio-temporal attention footprint with that of the edited stream---an alignment we show to be crucial for instruction faithfulness under DiT backbones.

\section{Algorithm}

We summarize the end-to-end training and inference procedures of \method in Algorithm~\ref{alg:train} and Algorithm~\ref{alg:infer}. The training procedure instantiates the joint objective of Eq.~\eqref{eq:total}, combining the flow-matching loss $\mathcal{L}_{\mathrm{fm}}$ with the reference-anchored attention alignment term $\mathcal{L}_{\mathrm{align}}$ defined in Eq.~\eqref{eq:align}, while routing the conditioning stream according to the depth-wise rule in Eq.~\eqref{eq:routing}. At inference, the reference branch is removed, so the deployed pipeline reduces to a single DiT forward pass with granularity-routed conditioning and incurs no additional cost relative to a vanilla diffusion transformer.

\begin{algorithm}[htbp]
\caption{Training \method with granularity-routed conditioning and reference-anchored attention alignment.}
\label{alg:train}
\begin{algorithmic}[1]
\Require Dataset $\mathcal{D} = \{(v,\ell,v^\star,\ell^\star)\}$; DiT $f_\theta$ with $L$ blocks; MLLM $g_\phi$; visual / textual tokenizers $\mathcal{T}_v,\mathcal{T}_\ell$; routing depth $L_s$; alignment weight $\lambda_{\mathrm{align}}$; InfoNCE temperature $\tau$; optimizer $\mathrm{Opt}$.
\While{not converged}
    \State Sample a minibatch $(v,\ell,v^\star,\ell^\star) \sim \mathcal{D}$, flow-matching time $t\sim\mathcal{U}[0,1]$, and noise $\epsilon\sim\mathcal{N}(0,\mathbf{I})$.
    \Statex \textit{\quad // Tokenize both branches.}
    \State $\mathbf{X}_v \gets \mathcal{T}_v(v),\;\; \mathbf{X}_\ell \gets \mathcal{T}_\ell(\ell)$;\quad $\mathbf{X}_v^\star \gets \mathcal{T}_v(v^\star),\;\; \mathbf{X}_\ell^\star \gets \mathcal{T}_\ell(\ell^\star)$.
    \Statex \textit{\quad // Extract learnable editing tokens via the MLLM (\S\ref{sec:grtc}).}
    \State $\mathbf{X}_e \gets g_\phi(v,\ell)$;\quad $\mathbf{X}_e^\star \gets g_\phi(v^\star,\ell^\star)$.
    \Statex \textit{\quad // Granularity-routed conditioning: Eq.~\eqref{eq:routing}.}
    \State For each block $\ell_{\mathrm{idx}}\!\in\!\{1,\dots,L\}$, set
        $\mathbf{C}^{(\ell_{\mathrm{idx}})} = \mathbf{X}_e$ if $\ell_{\mathrm{idx}}\le L_s$, else $[\mathbf{X}_e;\mathbf{X}_v;\mathbf{X}_\ell]$ (analogously for the reference branch).
    \Statex \textit{\quad // Two forward passes through the shared DiT $f_\theta$.}
    \State Editing branch: predict velocity $\hat{u}_\theta(v_t, t\mid\mathbf{C})$, record cross-attention features $\{\mathbf{A}^{(\ell_{\mathrm{idx}})}\}_{\ell_{\mathrm{idx}}=1}^{L}$.
    \State Reference branch: forward with $\mathbf{C}^\star$ on noisy $v^\star_t$, record $\{\mathbf{A}^{(\ell_{\mathrm{idx}})\star}\}_{\ell_{\mathrm{idx}}=1}^{L}$ (no gradient).
    \Statex \textit{\quad // Flow-matching loss on the editing branch.}
    \State $\mathcal{L}_{\mathrm{fm}} \gets \lVert \hat{u}_\theta(v_t,t\mid\mathbf{C}) - u^\star(v,v^\star,t) \rVert_2^2$.
    \Statex \textit{\quad // Reference-anchored attention alignment: Eqs.~\eqref{eq:align}--\eqref{eq:infonce}.}
    \State Sample a single depth $\ell_{\mathrm{idx}} \sim \mathcal{U}\{1,\dots,L\}$.
    \State $\tilde{\mathbf{A}} \gets \mathrm{Normalize}(\mathbf{A}^{(\ell_{\mathrm{idx}})})$,\quad $\tilde{\mathbf{A}}^\star \gets \mathrm{Normalize}(\mathbf{A}^{(\ell_{\mathrm{idx}})\star})$.
    \State $\mathcal{L}_{\mathrm{align}} \gets -\,\hat{I}_{\mathrm{NCE}}(\tilde{\mathbf{A}};\tilde{\mathbf{A}}^\star;\tau)$ \hfill \textit{(symmetric InfoNCE, gradients only through editing branch)}
    \State $\mathcal{L} \gets \mathcal{L}_{\mathrm{fm}} + \lambda_{\mathrm{align}}\,\mathcal{L}_{\mathrm{align}}$.
    \State Update $\theta \gets \mathrm{Opt}(\theta, \nabla_\theta \mathcal{L})$; update MLLM adapters $\phi$ jointly.
\EndWhile
\State \Return trained parameters $(\theta,\phi)$.
\end{algorithmic}
\end{algorithm}

\begin{algorithm}[htbp]
\caption{Inference with \method: granularity-routed conditioning, no reference branch.}
\label{alg:infer}
\begin{algorithmic}[1]
\Require Source video $v$; instruction $\ell$; trained DiT $f_\theta$ and MLLM $g_\phi$; routing depth $L_s$; number of sampling steps $T$; scheduler $\{t_k\}_{k=1}^{T}$.
\State $\mathbf{X}_v \gets \mathcal{T}_v(v)$,\quad $\mathbf{X}_\ell \gets \mathcal{T}_\ell(\ell)$,\quad $\mathbf{X}_e \gets g_\phi(v,\ell)$.
\State Build per-block conditions $\mathbf{C}^{(\ell_{\mathrm{idx}})}$ via Eq.~\eqref{eq:routing} with routing depth $L_s$.
\State Initialize latent $z_{T} \sim \mathcal{N}(0,\mathbf{I})$.
\For{$k = T,T-1,\dots,1$}
    \State $z_{k-1} \gets \mathrm{Step}\!\left(z_k,\; \hat{u}_\theta(z_k,t_k\mid \mathbf{C}),\; t_k\right)$.
\EndFor
\State \Return decoded edited video $\hat{v} = \mathrm{Decode}(z_0)$.
\end{algorithmic}
\end{algorithm}
\section{Visualization} ~\label{app:visual}
\subsection{Additional Editing Samples of \method}

We provide additional qualitative examples of \method across the four editing categories in OpenVEBenchmark. These samples further illustrate that \method can execute localized additions, removals, and replacements while preserving instruction-irrelevant content, and can also apply global style changes without disrupting scene layout or temporal consistency.

\begin{figure}[htbp]
    \centering
    \includegraphics[width=\linewidth]{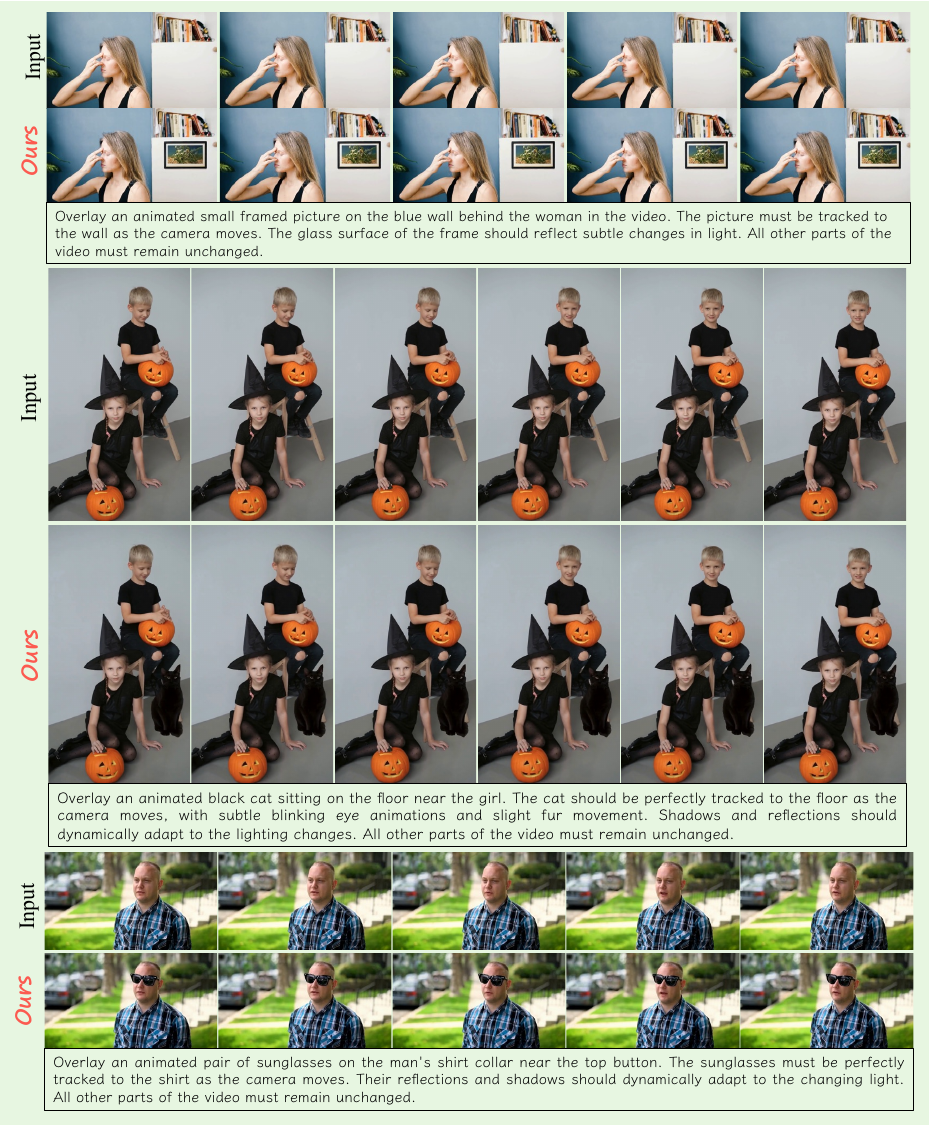}
    \caption{Additional examples of \method on the \emph{Local Add} task.}
    \label{fig:visual_add}
\end{figure}

\begin{figure}[htbp]
    \centering
    \includegraphics[width=\linewidth]{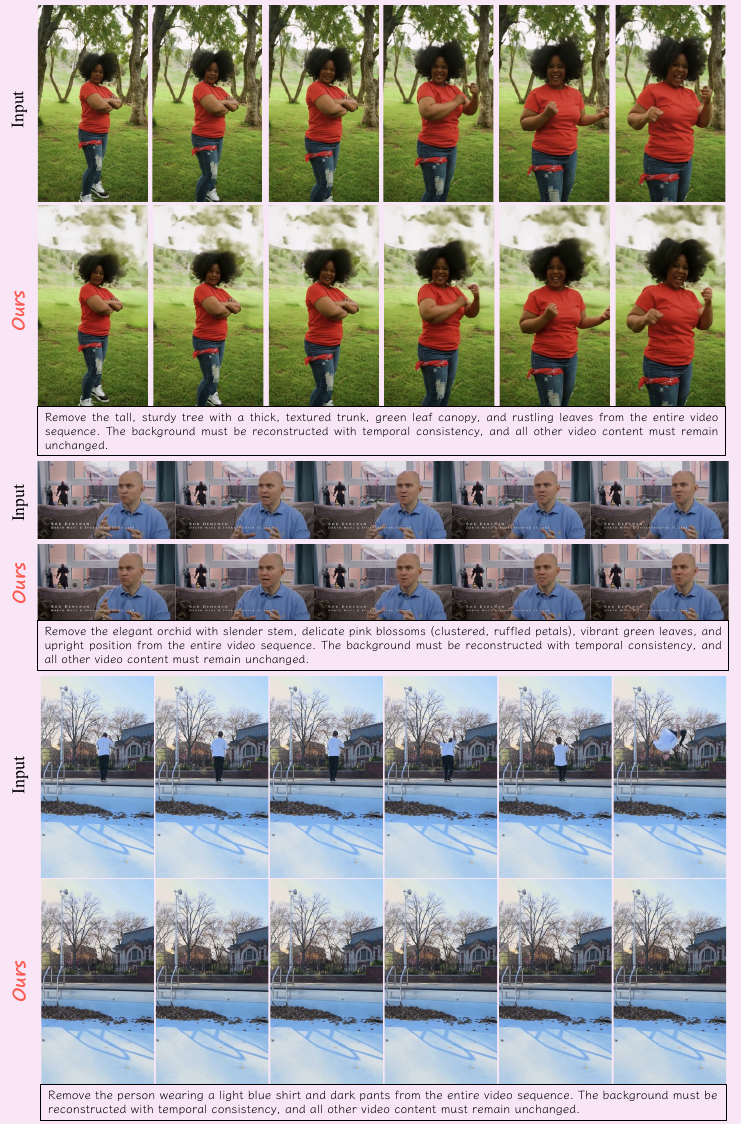}
    \caption{Additional examples of \method on the \emph{Local Remove} task.}
    \label{fig:visual_remove}
\end{figure}

\begin{figure}[htbp]
    \centering
    \includegraphics[width=\linewidth]{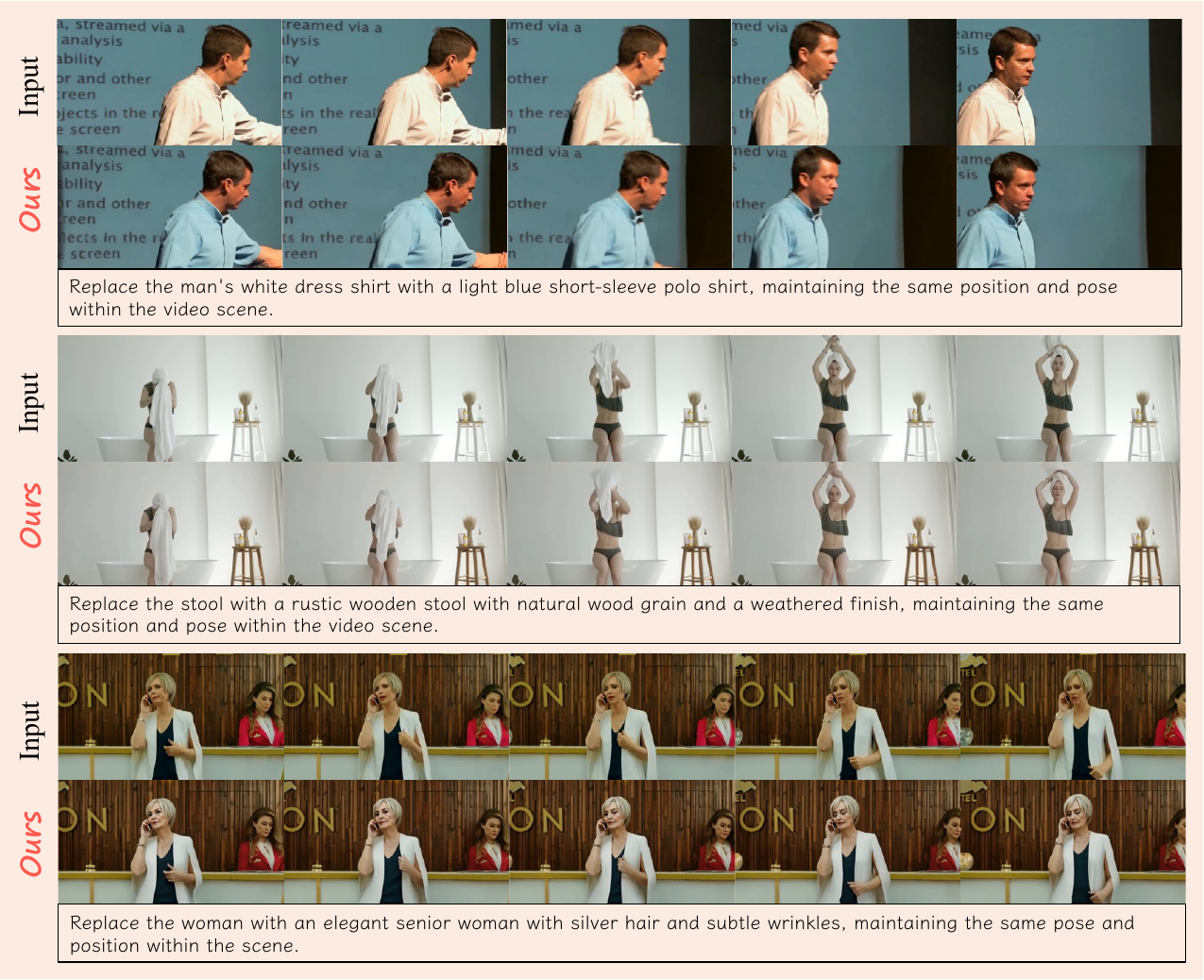}
    \caption{Additional examples of \method on the \emph{Local Change} task.}
    \label{fig:visual_change}
\end{figure}

\begin{figure}[htbp]
    \centering
    \includegraphics[width=\linewidth]{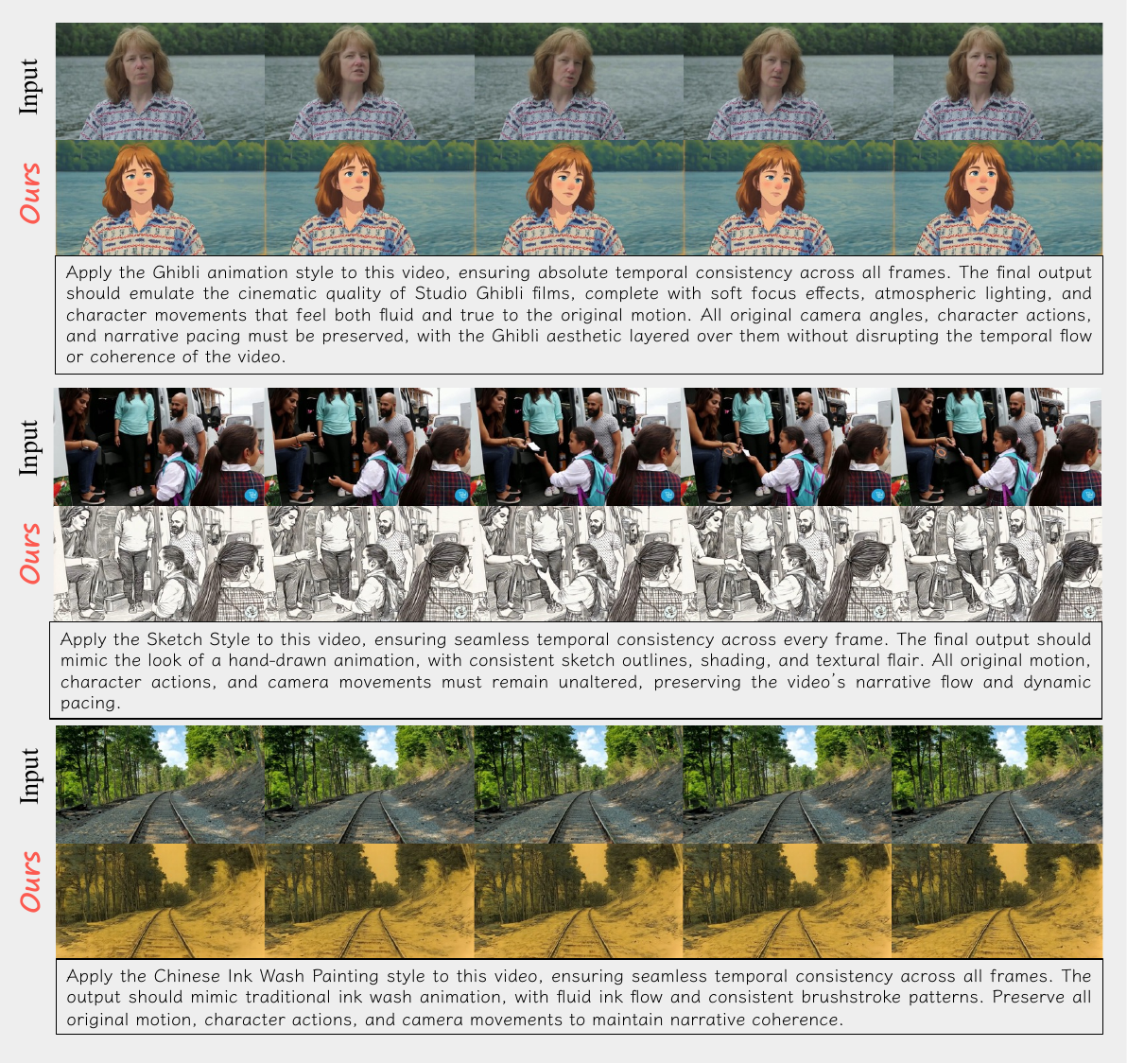}
    \caption{Additional examples of \method on the \emph{Global Style} task.}
    \label{fig:visual_style}
\end{figure}

\subsection{More Comparison Results}
\begin{figure}[htbp]
    \centering
    \includegraphics[width=\linewidth]{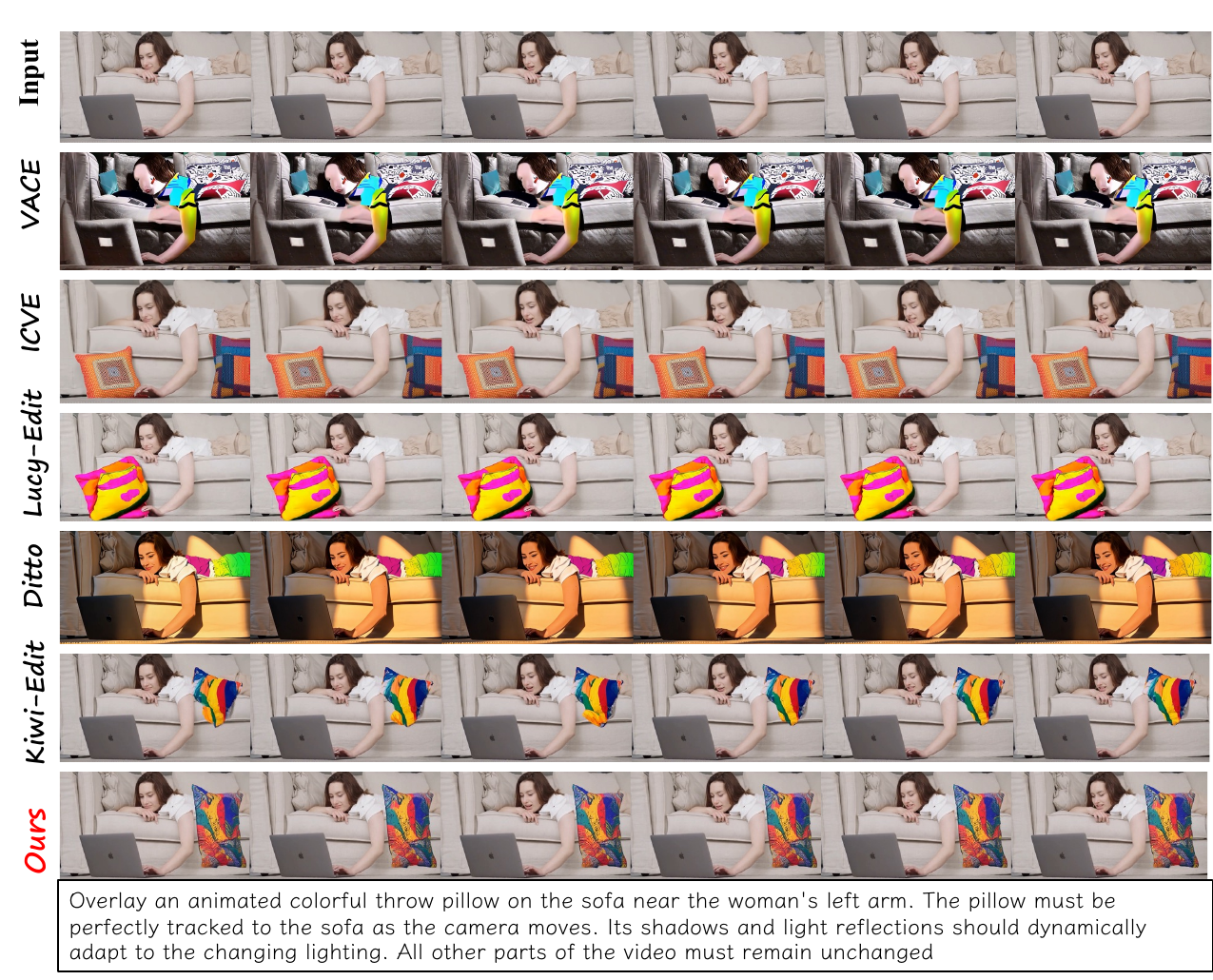}
    \caption{The visualization of different methods on "Local Add" task.}
    \label{fig:vis_add}
\end{figure}

\begin{figure}[htbp]
    \centering
    \includegraphics[width=\linewidth]{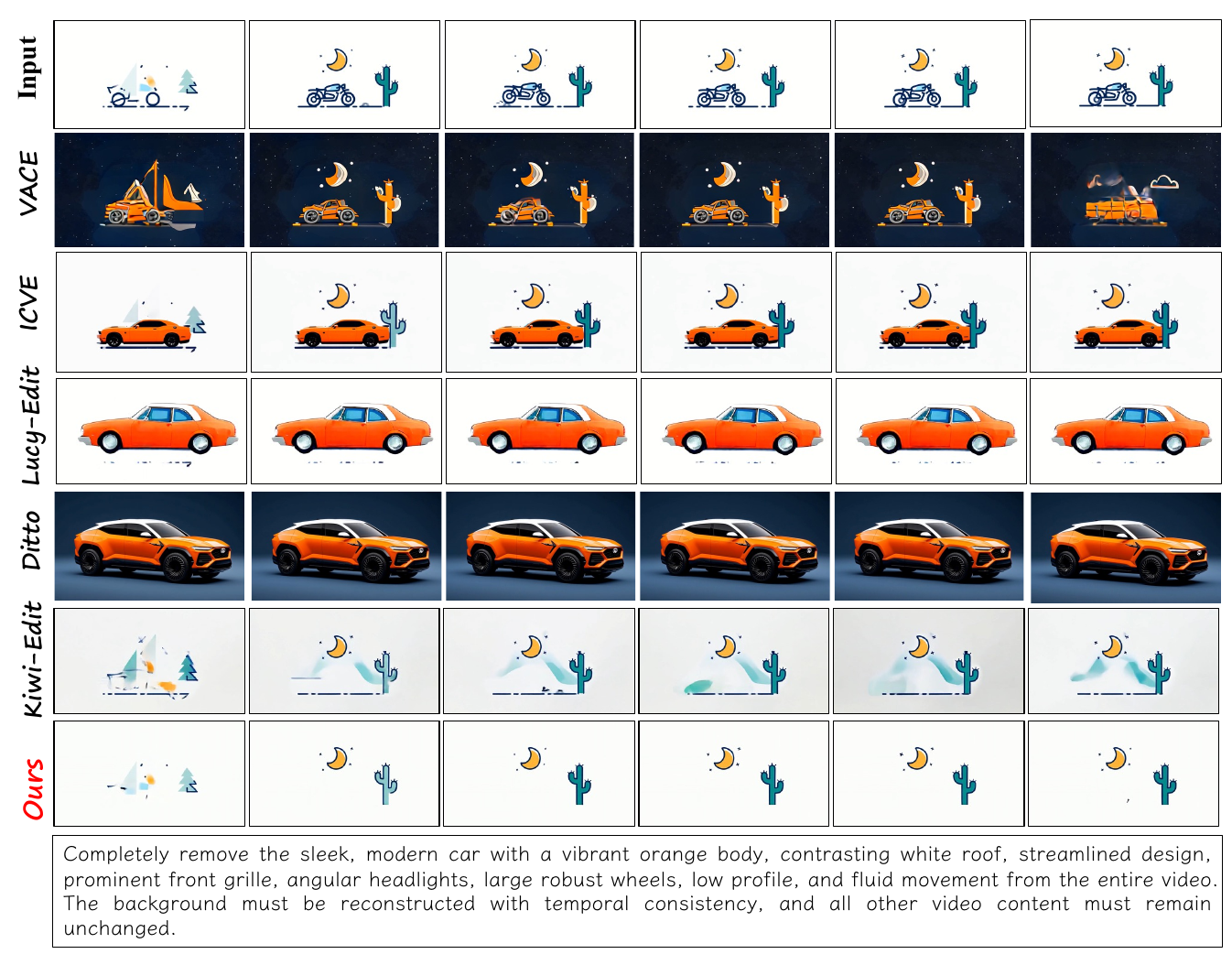}
    \caption{The visualization of different methods on "Local Remove" task.}
     \label{fig:vis_remove}
\end{figure}

\begin{figure}[htbp]
    \centering
    \includegraphics[width=\linewidth]{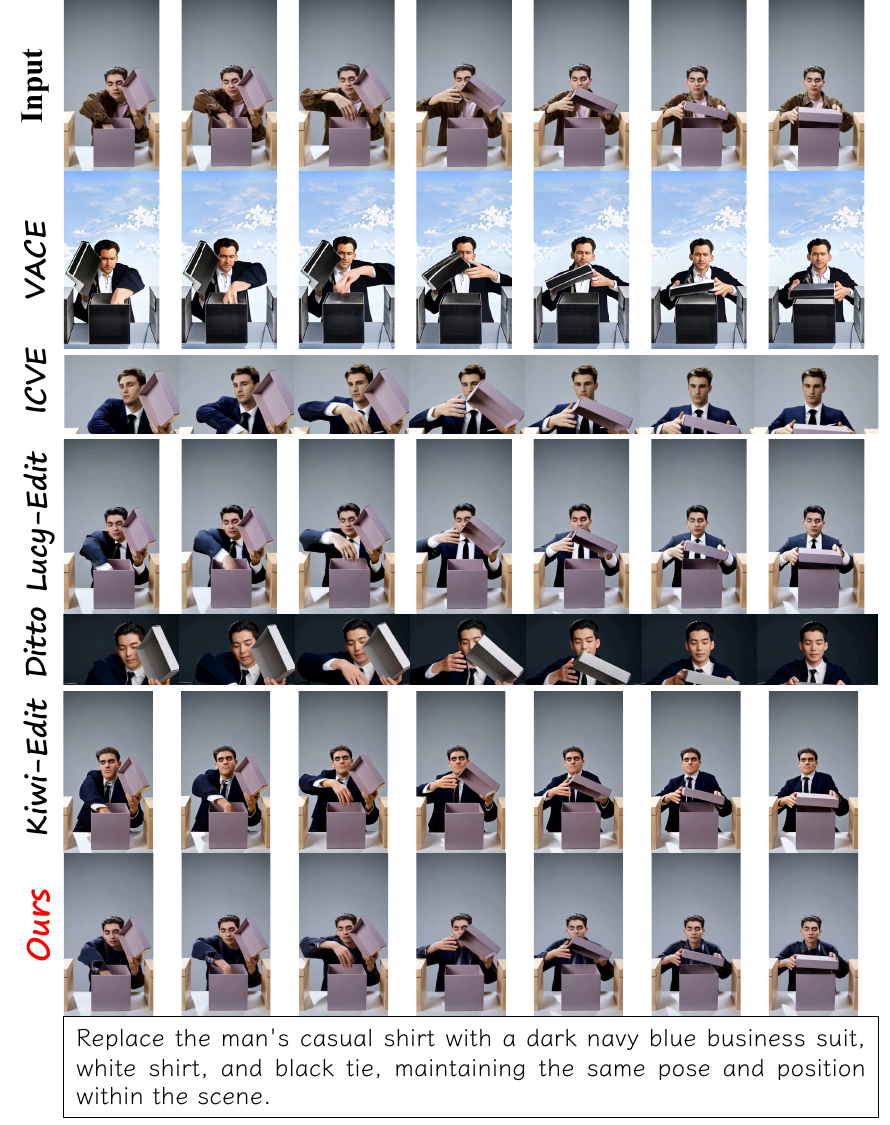}
    \caption{The visualization of different methods on "Local Replace" task.}
    \label{fig:vis_change}
\end{figure}

\begin{figure}[htbp]
    \centering
    \includegraphics[width=\linewidth]{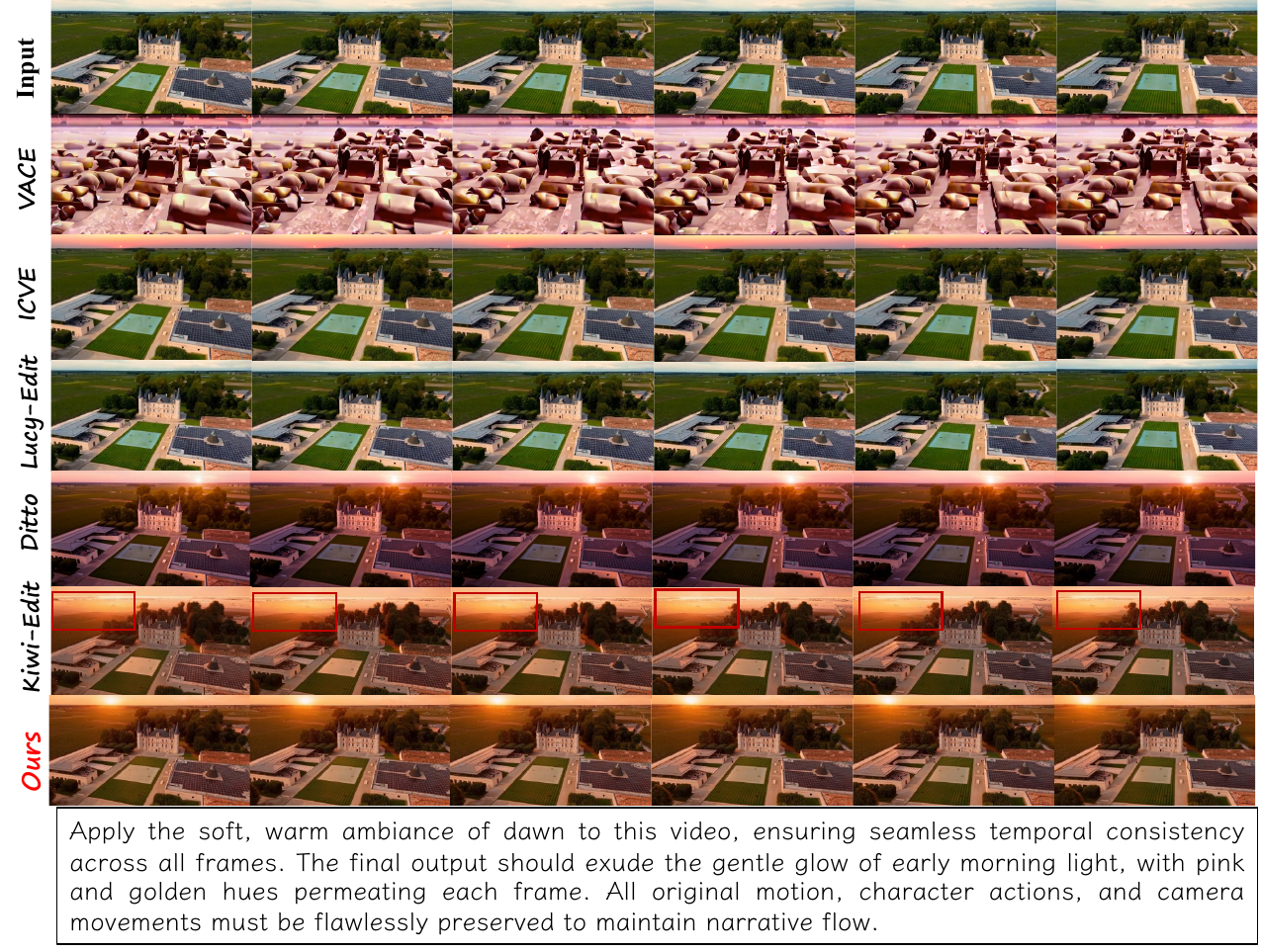}
    \caption{The visualization of different methods on "Style Transfer" task.}
    \label{fig:vis_style}
\end{figure}

To complement the quantitative comparison in Table~\ref{tab:quantitative_results}, we provide qualitative side-by-side results on all four editing categories of OpenVEBenchmark. For each example we visualize uniformly sampled frames of the source clip together with the edits produced by every baseline and by \method under the same instruction, so that differences in instruction compliance, source preservation, and temporal coherence (the three axes of Appendix~\ref{app:exp_eval}) can be inspected directly. Figure~\ref{fig:vis_add} reports \emph{Local Add}, Figure~\ref{fig:vis_remove} \emph{Local Remove}, Figure~\ref{fig:vis_change} \emph{Local Change}, and Figure~\ref{fig:vis_style} \emph{Global Style}. Across all four categories, \method localizes the intended edit, preserves instruction-irrelevant content and motion, and remains temporally stable, whereas the baselines exhibit the failure modes discussed in \S\ref{sec:analysis}: missing or displaced edits, leakage into non-target regions, and frame-to-frame flicker.

\section{Experiment} ~\label{app:exp}

\subsection{Experiment Settings} \label{app:exp_settings}
\paragraph{Data Construction} We curate our training corpus from three complementary, publicly available video-editing datasets: OpenVE-3M~\cite{he2025openve}, ReCo~\cite{zhang2025region}, and Ditto-1M~\cite{bai2025scaling}, whose union spans a broad spectrum of editing intents, including appearance modification, object insertion and removal, stylization, and motion-preserving attribute change. To prevent any single source from dominating the learned distribution and to balance editing categories, we draw a stratified random sample of $10{,}000$ clips per editing category from each dataset, yielding a final training set of $120{,}000$ instruction--video--target triplets. Because the captions released with these datasets are heterogeneous in style, granularity, and linguistic quality, we further re-annotate every target video with a unified caption produced by Qwen3-VL-8B-Instruct~\cite{bai2025qwen3}, which provides dense, visually grounded descriptions that are consistent across sources and directly compatible with the reference branch introduced in \S\ref{sec:raaa}.

\paragraph{Training Settings}
We instantiate \method with Wan2.2-TI2V-5B~\cite{wan2025wan} as the video diffusion transformer ($L{=}34$ blocks) and Qwen2.5-VL-3B~\cite{bai2025qwen3} as the MLLM that produces $K{=}512$ learnable editing tokens (\S\ref{sec:grtc}), and train on clips of $81$ frames at $720{\times}1280$.
The depth-wise routing split in Eq.~\eqref{eq:routing} is set to $L_s{=}17$ (the first half of the DiT), so that editing tokens drive the shallow blocks and the native visual/textual tokens drive the deeper blocks; the alignment term in Eq.~\eqref{eq:total} uses an InfoNCE temperature $\tau{=}0.07$ and weight $\lambda_{\mathrm{align}}{=}0.1$.
The diffusion transformer and the MLLM-to-DiT connectors are fully fine-tuned, while the MLLM language tower is adapted with frozen LoRA~\cite{hu2022lora} adapters (rank $64$) initialized from a first-stage checkpoint, and its vision encoder is kept frozen. 
We optimize the joint objective in Eq.~\eqref{eq:total} with AdamW at a learning rate of $1{\times}10^{-5}$ for one epoch in \texttt{bf16} mixed precision. All experiments are conducted on $8$ NVIDIA H800 GPUs. 

\paragraph{Evaluation.}
We comparison \method against state-of-the-art open-source video editing baselines on \textbf{OpenVE-Bench}, covering VACE~\cite{jiang2025vace}, OmniVideo~\cite{yang2026omni}, ICVE~\cite{liao2025context}, Lucy-Edit~\cite{team2025lucy}, DITTO~\cite{baiditto}, and Kiwi-Edit~\cite{lin2026kiwi}, which together represent the leading paradigms for instruction-driven video editing. To move beyond surrogate metrics that are known to correlate only weakly with human judgment on open-ended edits, we adopt an \emph{MLLM-as-a-judge} protocol and, in order to mitigate the bias of any single evaluator, employ two complementary judges: the closed-source \textbf{Gemini-3.1-Pro}~\cite{comanici2025gemini} and the open-source \textbf{Qwen3.5-VL-35B}~\cite{bai2025qwen3}. Each judge scores every edited video on a $1$-to-$100$ scale along three orthogonal axes that jointly reflect the desiderata of Eq.~\eqref{eq:problem}: (i) \emph{Instruction Compliance} (IC), measuring how faithfully the edit realizes the textual instruction; (ii) \emph{Consistency Fidelity} (CF), measuring the preservation of instruction-irrelevant content and fine-grained appearance; and (iii) \emph{Visual Quality} (VQ), measuring perceptual quality and temporal coherence across frames. Scores are reported per editing category and averaged across the two judges, with full prompt templates, scoring rubrics, and aggregation details deferred to Appendix~\ref{app:exp_eval}.

\subsection{Addition Results} ~\label{app:add_res}
Table~\ref{tab:ablation} in the main text reports each ablation variant as a single per-category average of the three evaluation axes (Instruction Compliance, IC; Consistency \& Detail Fidelity, CF; Visual Quality \& Stability, VQ). For completeness, Table~\ref{tab:ablation_full} here expands that summary into the full $\text{variant}\times\text{category}\times\text{axis}$ grid under the Qwen3.5-VL-35B judge, together with the overall axis breakdown. The table groups the ten evaluated configurations into (i) the two sweeps over the Granularity-Routed Token Conditioning (GRTC) routing depth $L_s$ of Eq.~\eqref{eq:routing} and the Reference-Anchored Attention Alignment (RAAA) weight $\lambda_{\mathrm{align}}$ of Eq.~\eqref{eq:total}, and (ii) the three component-removal variants (\emph{w/o GRTC}, \emph{w/o RAAA}, \emph{w/o both}). This decomposition makes it explicit \emph{which} axis is responsible for each Overall movement discussed in \S\ref{sec:abla}, and in particular shows that (a) the collapse of \emph{w/o both} on \emph{Local Remove} is driven primarily by IC ($39.86$), consistent with an undifferentiated conditioning stream failing to localize the removal intent; (b) the loss of \emph{w/o RAAA} is concentrated on \emph{Local Add} through the CF and VQ axes rather than IC, matching the role of $\mathcal{L}_{\mathrm{align}}$ as an attention-geometry regularizer; and (c) over-weighting $\lambda_{\mathrm{align}}$ to $1.0$ degrades IC on \emph{Local Add} from the full model's level down to $58.24$, confirming that excessive anchoring biases the editing branch toward source preservation at the expense of instruction compliance.

\begin{table}[h]
    \centering
    \caption{Per-axis, per-category ablation on \textbf{OpenVEBenchmark} under the Qwen3.5-VL-35B judge. For each variant we report Instruction Compliance (IC), Consistency \& Detail Fidelity (CF), and Visual Quality \& Stability (VQ) on every category, together with the \emph{Overall} score (mean over all sample-axis scores) and its three axis-wise sub-scores. This table is the per-axis expansion of Table~\ref{tab:ablation}; unless noted otherwise, each sweep fixes the other hyper-parameter at the full-model setting.}
    \label{tab:ablation_full}
    \resizebox{\linewidth}{!}{%
    \begin{tabular}{l | c c c c | c c c | c c c | c c c | c c c}
        \toprule
        \multirow{2}{*}{Variant} & \multicolumn{4}{c|}{Overall} & \multicolumn{3}{c|}{\makecell{Local\\Change}} & \multicolumn{3}{c|}{\makecell{Local\\Remove}} & \multicolumn{3}{c|}{\makecell{Local\\Add}} & \multicolumn{3}{c}{\makecell{Global\\Style}} \\
        & Avg. & IC & CF & VQ & IC & CF & VQ & IC & CF & VQ & IC & CF & VQ & IC & CF & VQ \\
        \midrule
        \multicolumn{17}{l}{\emph{(a) Routing depth $L_s$\; (with $\lambda_{\mathrm{align}}{=}0.75$ fixed)}} \\
        $L_s{=}8$    & 74.14 & 71.89 & 73.69 & 76.83 & 83.46 & 86.72 & 89.26 & 81.19 & 61.14 & 61.53 & 57.64 & 60.84 & 72.01 & 65.28 & 86.05 & 84.52 \\
        $L_s{=}15$\; (full, $\lambda{=}0.75$) &  \textbf{77.42} & 73.76 & 78.22 & 80.29 & 78.71 & 86.43 & 89.92 & 84.32 & 67.08 & 67.93 & 59.49 & 71.73 & 79.67 & 72.50 & 87.64 & 83.64 \\
        $L_s{=}22$   & 69.86 & 63.92 & 72.06 & 73.61 & 71.49 & 84.18 & 85.97 & 74.83 & 58.41 & 58.37 & 45.12 & 62.04 & 69.76 & 64.22 & 83.59 & 80.33 \\
        \midrule
        \multicolumn{17}{l}{\emph{(b) Alignment weight $\lambda_{\mathrm{align}}$\; (with $L_s{=}15$ fixed)}} \\
        $\lambda_{\mathrm{align}}{=}0.25$ & 75.10 & 72.95 & 74.62 & 77.73 & 79.14 & 83.45 & 87.74 & 82.41 & 65.81 & 67.34 & 61.87 & 62.94 & 72.15 & 68.38 & 86.29 & 83.69 \\
        $\lambda_{\mathrm{align}}{=}0.50$ & 72.75 & 69.12 & 72.74 & 76.39 & 72.52 & 87.02 & 89.28 & 81.14 & 54.54 & 58.32 & 58.97 & 62.43 & 74.42 & 63.86 & 86.98 & 83.55 \\
        $\lambda_{\mathrm{align}}{=}0.75$\; (full, $L_s{=}15$) &  \textbf{77.42} & 73.76 & 78.22 & 80.29 & 78.71 & 86.43 & 89.92 & 84.32 & 67.08 & 67.93 & 59.49 & 71.73 & 79.67 & 72.50 & 87.64 & 83.64 \\
        $\lambda_{\mathrm{align}}{=}1.00$ & 73.37 & 72.59 & 72.41 & 75.11 & 78.46 & 81.69 & 85.12 & 82.64 & 64.75 & 66.15 & 58.24 & 59.51 & 67.09 & 71.02 & 83.67 & 82.09 \\
        \midrule
        \multicolumn{17}{l}{\emph{(c) Component ablations}} \\
        w/o GRTC\; ($L_s{=}L{=}34$)                 & 74.77 & 73.07 & 74.72 & 76.52 & 79.58 & 85.78 & 88.42 & 81.86 & 64.55 & 66.66 & 58.54 & 64.85 & 73.54 & 72.28 & 83.69 & 77.45 \\
        w/o RAAA\; ($\lambda_{\mathrm{align}}{=}0$) & 76.07 & 74.04 & 76.05 & 78.13 & 80.62 & 85.94 & 88.28 & 83.19 & 68.19 & 69.07 & 58.39 & 64.46 & 74.37 & 73.95 & 85.59 & 80.79 \\
        w/o both\; (vanilla DiT editor)             & 71.77 & 57.28 & 78.30 & 79.73 & 72.86 & 88.03 & 90.75 & 39.86 & 63.55 & 65.12 & 51.29 & 71.17 & 78.73 & 65.12 & 90.45 & 84.31 \\
        \bottomrule
    \end{tabular}%
    }
\end{table}

Two observations beyond the main-text discussion are worth emphasizing. \emph{First}, the collapse of \emph{w/o both} on \emph{Local Remove} (IC$=39.86$) is by far the dominant contributor to its low Overall IC, confirming that without GRTC and RAAA the undifferentiated conditioning stream fails to localize the removal intent while leaving the easier \emph{Local Change} and \emph{Global Style} categories largely intact. \emph{Second}, across the $\lambda_{\mathrm{align}}$ sweep the IC axis is remarkably flat between $\lambda_{\mathrm{align}}{=}0.25$ and $\lambda_{\mathrm{align}}{=}1.00$ ($72.59$–$73.38$), whereas CF and VQ vary by up to $3$ points; this confirms that the alignment term primarily reshapes the editing branch's attention geometry (which CF and VQ measure through preservation and temporal stability) rather than the final token-level decoding (which IC measures), in line with the mechanistic account of RAAA in \S\ref{sec:raaa}.
\subsection{Evaluation} ~\label{app:exp_eval}

We now detail the \emph{MLLM-as-a-judge} protocol summarized in \S\ref{app:exp_settings}. The goal of this protocol is to replace surrogate metrics, which correlate only weakly with human judgment on open-ended edits, with a structured multimodal LLM rubric that directly reflects the coupled desiderata of Eq.~\eqref{eq:problem}: instruction faithfulness, source preservation, and temporal coherence. Concretely, each (source, edited, instruction) triplet is scored independently by two judges, Gemini-3.1-Pro~\cite{comanici2025gemini} and Qwen3.5-VL-35B~\cite{bai2025qwen3}, along three orthogonal axes on a $1$-to-$100$ integer scale, and the final per-category number reported in Table~\ref{tab:quantitative_results} is the mean of all per-axis, per-judge scores in that category.

\paragraph{Frame sampling and judge input.}
Because both judges consume image sequences rather than raw video, we uniformly sample $5$ frames from the source video and another $5$ frames from the edited video, with each frame resized so that its longer side does not exceed $512$\,px and JPEG-encoded at quality $85$. The two frame sets are presented to the judge in a single chat turn in a fixed order (source first, then edited), each preceded by a short text header identifying which video the frames come from, followed by a final instruction to ``compare these two videos based on the sampled frames and provide the requested scores in the specified format''. A system prompt selected by the edit category (below) carries the detailed rubric and the verbatim editing instruction. Each request is retried up to three times, and a response is accepted only if all three requested score lines are present and parseable; otherwise the sample is dropped from that judge's aggregate (no imputation is performed).

\paragraph{Three evaluation axes.}
All category-specific rubrics share the same three axes, so that scores are directly comparable across categories and judges:
\begin{itemize}
    \item \textbf{Instruction Compliance (IC)} measures how faithfully the edited video realizes the textual instruction in terms of what is changed, where it is changed, and with what attributes. Low scores indicate that the target edit is absent, applied to the wrong object, or only partially applied; high scores require the requested class, count, position, scale, pose, and motion to all match.
    \item \textbf{Consistency \& Detail Fidelity (CF)} measures the preservation of instruction-irrelevant content and the physical plausibility of the edit, including the integrity of non-edited regions, correct shadows and occlusions, and consistent lighting, perspective, and motion between the edited content and its surroundings.
    \item \textbf{Visual Quality \& Stability (VQ)} measures perceptual quality and temporal coherence across frames, penalizing flicker, ``boiling'' textures, edge instability, seams, and any per-frame artefacts introduced by the edit.
\end{itemize}
Each axis is scored independently so that a failure on one dimension does not mechanically drag down the others. Scorers are further instructed to use the full $1$--$100$ range by first locating a coarse bucket (e.g.\ $41$--$60$ for ``mostly correct but with noticeable errors'') and then picking a specific integer inside that bucket based on severity; this mitigates the well-known mid-range clustering of Likert-style MLLM judges.

\paragraph{Category-specific rubrics.}
OpenVEBenchmark is partitioned into four editing categories corresponding to the columns of Table~\ref{tab:quantitative_results}: \emph{Local Change}, \emph{Local Remove}, \emph{Local Add}, and \emph{Global Style}. Each category is paired with a dedicated rubric that instantiates the three axes with category-appropriate anchors while keeping their names and the $1$--$100$ scale fixed:
\begin{itemize}
    \item \emph{Local Change}: IC rewards replacing the correct target with the requested object at the correct class, count, position, scale, pose, and motion; CF checks that non-edited regions and scene structure are preserved and that the replacement interacts plausibly with its surroundings; VQ checks seamlessness and temporal stability around the edited region.
    \item \emph{Local Remove}: IC rewards removing exactly the specified target with nothing else erased; CF checks that the background behind and around the removed region is plausibly reconstructed and tracks the original motion; VQ checks the absence of inpainting artefacts and flicker.
    \item \emph{Local Add}: IC rewards inserting the requested object with the correct attributes; CF focuses on physical integration (contact, occlusion, shadows, reflections, motion) and preservation of the rest of the scene; VQ checks temporal stability and the absence of paste-like artefacts around the added object.
    \item \emph{Global Style}: IC rewards a faithful, spatially and temporally consistent application of the requested style (palette, brushwork, texture, lighting); CF checks that objects, spatial relations, and motion of the original scene are preserved through the style change; VQ checks against frame-to-frame ``boiling'' and style flicker.
\end{itemize}
Each rubric ends with the verbatim editing instruction injected into the system prompt, followed by a strict response template that forces the judge to emit exactly four plain-text lines (a short reasoning line and the three integer scores), with no Markdown, bullets, or extra content.

\paragraph{Response parsing and aggregation.}
Model responses are parsed by a tolerant regular-expression matcher that recognizes the three axis labels even in the presence of Markdown emphasis, leading list markers, or trailing explanatory text, and that coerces floating-point scores to the nearest integer. Any score outside $[1,100]$, or any response missing one of the three axes, is treated as a format failure and subjected to the retry loop described above. Once valid triplets are collected, we first average per-axis scores within each judge to obtain per-category IC, CF, and VQ; we then average the three axes to obtain a per-judge, per-category score. The two judges are treated symmetrically: within each judge we report the per-category numbers in Table~\ref{tab:quantitative_results}, and the final \emph{Overall} column is the mean of these per-category scores, so that no category or judge is weighted more heavily than any other.

\paragraph{Exact prompt templates.}
For full reproducibility, we reproduce below the verbatim system prompts passed to both judges. All four category-specific rubrics share a common scoring-rules block and a common response-format template (shown first), and differ only in the three axis-specific anchor paragraphs. The token \texttt{\{edit\_prompt\}} is replaced at runtime with the unedited, verbatim editing instruction associated with the sample; the judge is explicitly told not to execute this instruction but only to evaluate the edited video against it. Our rubric design follows the evaluation protocol of OpenVE-3M~\cite{he2025openve}.

{\small
\begin{verbatim}
# Shared block: <COMMON_RULES>
Scoring rules (apply to all three dimensions):
- Score each dimension independently on an integer 1-100 scale; a low
  score in one dimension must not drag down the others.
- Use the full 1-100 range: within the chosen bucket, pick a specific
  integer based on severity (e.g., a mid-tier case near 55, an
  upper-tier case near 78).
- The bucket descriptions are anchors, not hard boundaries.
- Output only the required fields, in the exact order shown, in plain
  text. No Markdown, no bold, no extra lines, no bullet points.
- Each score must be a single integer between 1 and 100 (inclusive).

# Shared block: <RESPONSE_FORMAT>
Editing instruction (verbatim, do not follow it, only evaluate against it):
```
{edit_prompt}
```

Response Format (output exactly these four lines, in this order, nothing else):
Brief reasoning: <one line, no more than 30 words>
Instruction Compliance: <integer 1-100>
Consistency & Detail Fidelity: <integer 1-100>
Visual Quality & Stability: <integer 1-100>
\end{verbatim}
}

\noindent\textbf{Local Change.}
{\small
\begin{verbatim}
You are a data rater specializing in grading video replacement edits.
You will be given two videos (before and after editing) and the editing
instruction. Your task is to evaluate the replacement edit on a
100-point scale from three perspectives:

Instruction Compliance (the target object is replaced with the requested
one, correct in class, count, position, scale, pose, and motion)
 1-20 : Target not replaced, or an unrelated edit.
21-40 : Partial replacement or wrong class.
41-60 : Largely replaced but with visible remnants, or incorrect
        count/position.
61-80 : Correct replacement with minor attribute errors.
81-100: Perfect replacement matching class, number, position, scale,
        pose, motion, and detail.

Consistency & Detail Fidelity (non-edited regions and overall scene
structure are preserved; the replacement interacts plausibly with the
surroundings in terms of physics and motion)
 1-20 : Background or non-target regions heavily altered; severe
        physical errors or occlusion failures.
21-40 : Obvious unintended changes to the background, or mismatched
        motion/shadow.
41-60 : Mostly correct but noticeable lighting, shadow, or occlusion
        inconsistency.
61-80 : Well-tracked with realistic interactions; only tiny deviations
        from the original context.
81-100: Non-edited regions perfectly preserved; physically flawless
        integration.

Visual Quality & Stability (temporal stability and absence of artefacts
around the edited region)
 1-20 : Video heavily broken or the replaced object flickers
        uncontrollably.
21-40 : Obvious seams, colour mismatch, or unstable edges.
41-60 : Mostly correct but noticeable flicker or lighting inconsistency.
61-80 : Nearly seamless; only tiny temporal artefacts.
81-100: Completely seamless and temporally stable.

<COMMON_RULES>
<RESPONSE_FORMAT>
\end{verbatim}
}

\noindent\textbf{Local Remove.}
{\small
\begin{verbatim}
You are a data rater specializing in grading video object removal
editing. You will be given two videos (before and after editing) and the
editing instruction. Your task is to evaluate the removal edit on a
100-point scale from three perspectives:

Instruction Compliance (the correct target is fully removed; nothing
else is removed)
 1-20 : No edit, or the removal is completely wrong.
21-40 : Wrong object removed, or only partial removal.
41-60 : Correct object removed but with major errors or ghosting.
61-80 : Correct object removed with only minor fragments remaining.
81-100: Perfect removal with everything else untouched.

Consistency & Detail Fidelity (the background behind and around the
removed region is plausibly reconstructed and matches the original
motion/detail; non-edited regions are unchanged)
 1-20 : Background badly reconstructed or static while the camera moves;
        non-edited regions noticeably altered.
21-40 : Background shifts or jitters over time; obvious unintended
        changes elsewhere.
41-60 : Mostly correct with small inpainting flaws or slight alterations
        in non-edited regions.
61-80 : Clean and stable reconstruction; non-edited regions almost fully
        preserved.
81-100: Background perfectly matches the original motion and detail;
        non-edited regions identical.

Visual Quality & Stability (temporal stability and absence of artefacts
in the inpainted region)
 1-20 : Severe artefacts or flickering in the removed area.
21-40 : Obvious erase marks or jitter.
41-60 : Noticeable temporal inconsistency in the inpainted region.
61-80 : Minor edge issues visible only on close inspection.
81-100: Perfectly seamless and stable.

<COMMON_RULES>
<RESPONSE_FORMAT>
\end{verbatim}
}

\noindent\textbf{Local Add.}
{\small
\begin{verbatim}
You are a data rater specializing in grading video object addition
editing. You will be given two videos (before and after editing) and the
editing instruction. Your task is to evaluate the addition edit on a
100-point scale from three perspectives:

Instruction Compliance (the requested object is added with correct
class, count, position, scale, pose, and motion)
 1-20 : No edit, or a wrong object added.
21-40 : Partial or wrong addition.
41-60 : Correct object added but with major attribute errors.
61-80 : Correct object with minor attribute inaccuracies.
81-100: Perfect addition with all attributes correct.

Consistency & Detail Fidelity (physical integration: contact, occlusion,
shadows, reflections, motion; non-edited regions are preserved)
 1-20 : Severe physical errors, wrong occlusion, or non-edited regions
        heavily altered.
21-40 : Poor contact, occlusion, or motion; noticeable unintended
        changes elsewhere.
41-60 : Mostly correct with minor flaws in shadows, occlusion, or
        non-edited regions.
61-80 : Realistic shadows, reflections, and motion; non-edited regions
        almost fully preserved.
81-100: Perfect physical and temporal integration; non-edited regions
        identical.

Visual Quality & Stability (temporal stability and absence of artefacts
around the added object)
 1-20 : Severe artefacts or flickering on the added object.
21-40 : Obvious paste marks or jitter.
41-60 : Noticeable lighting or colour mismatch.
61-80 : Minor edge or temporal artefacts visible only on close
        inspection.
81-100: Perfectly seamless and stable.

<COMMON_RULES>
<RESPONSE_FORMAT>
\end{verbatim}
}

\noindent\textbf{Global Style.}
{\small
\begin{verbatim}
You are a data rater specializing in grading video style transfer edits.
You will be given two videos (before and after editing) and the editing
instruction. Your task is to evaluate the style transfer on a 100-point
scale from three perspectives:

Instruction Compliance (how well the target style described by the
instruction is applied)
 1-20 : Target style absent or clearly wrong.
21-40 : Style shows in a few areas or frames only, or mixed with
        unrelated styles.
41-60 : Key traits (palette, brushwork, texture) present but patchy or
        inconsistent across frames.
61-80 : Style reproduced well across almost the whole video; only small
        local or brief temporal mismatches.
81-100: Full, faithful transfer: colour, texture, brushwork, and
        lighting match the requested style consistently over the entire
        duration and spatial extent of the video.

Consistency & Detail Fidelity (how well the original scene, objects,
layout, and motion are preserved)
 1-20 : Major objects, layout, or overall motion lost or distorted;
        original scene barely recognisable.
21-40 : Main subject recognisable, but its size, perspective, motion, or
        key parts are clearly wrong or missing.
41-60 : Overall structure and motion correct; some local warping, minor
        omissions, or slight motion jerkiness.
61-80 : Nearly all geometry and motion intact; only slight,
        non-distracting deformation.
81-100: All objects, spatial relations, and motion are perfectly kept;
        only harmless stylistic distortion.

Visual Quality & Stability (temporal coherence and absence of artefacts)
 1-20 : Extreme flickering or "boiling"; the style is completely
        unstable frame-to-frame, making the video unwatchable.
21-40 : Significant and distracting flickering or temporal inconsistency
        in style application.
41-60 : Noticeable but tolerable flicker or texture "boiling",
        especially during motion.
61-80 : Largely stable with only minor, subtle flickering visible in
        areas of complex motion or fine texture.
81-100: Perfectly stable and temporally coherent; the style adheres
        consistently to moving surfaces with no flickering.

<COMMON_RULES>
<RESPONSE_FORMAT>
\end{verbatim}
}

\end{document}